\documentclass{article} %
\usepackage{bvb_arxiv,times}

\usepackage{hyperref}
\usepackage{url}
\usepackage{graphicx}
\usepackage{booktabs}
\usepackage{amsmath,amssymb}
\usepackage{multirow}
\usepackage{algorithm}
\usepackage{algpseudocode}
\usepackage{xcolor}
\usepackage{colortbl}
\usepackage{tikz}
\usepackage{float}
\usepackage{rotating}
\usepackage{caption}
\usepackage{listings}

\definecolor{AxDV}{HTML}{0072B2}
\definecolor{AxLS}{HTML}{E14B72}
\definecolor{AxOvr}{HTML}{8B5CF6}
\newcommand{\DualVQA}{\textcolor{AxDV}{\textbf{Dual VQA}}}
\newcommand{\LatentSim}{\textcolor{AxLS}{\textbf{Latent Similarity}}}

\newcommand{\SqrtMean}{\textcolor{AxOvr}{\textbf{Overall}}}
\newcommand{\abDV}{\textcolor{AxDV}{\textbf{DV}}}
\newcommand{\abLS}{\textcolor{AxLS}{\textbf{LS}}}

\newcommand{\logoGPT}{\raisebox{-0.2\height}{\includegraphics[height=1.0em]{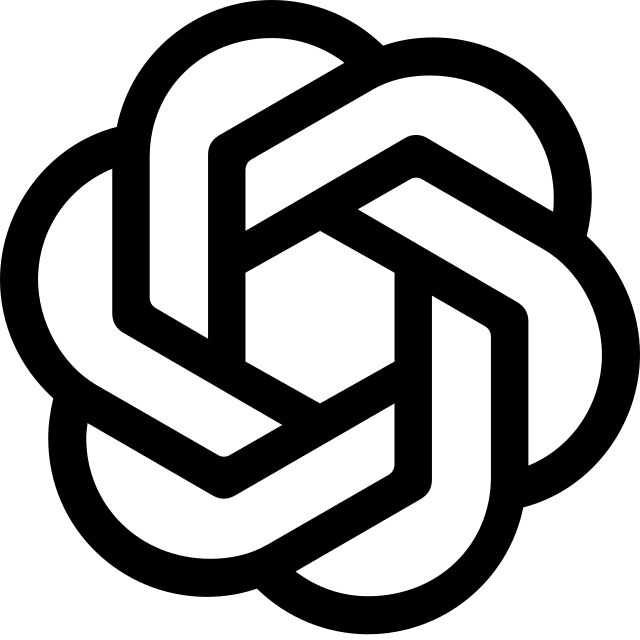}}}
\newcommand{\logoGrok}{\raisebox{-0.2\height}{\includegraphics[height=1.0em]{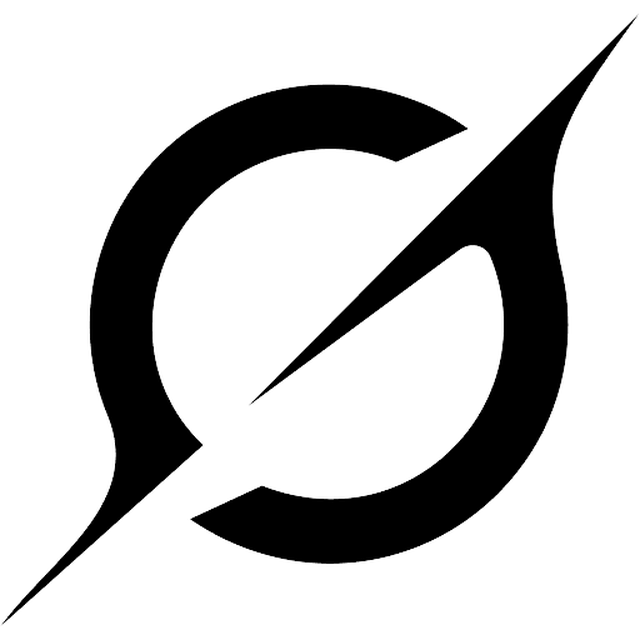}}}
\newcommand{\logoGemini}{\raisebox{-0.2\height}{\includegraphics[height=1.0em]{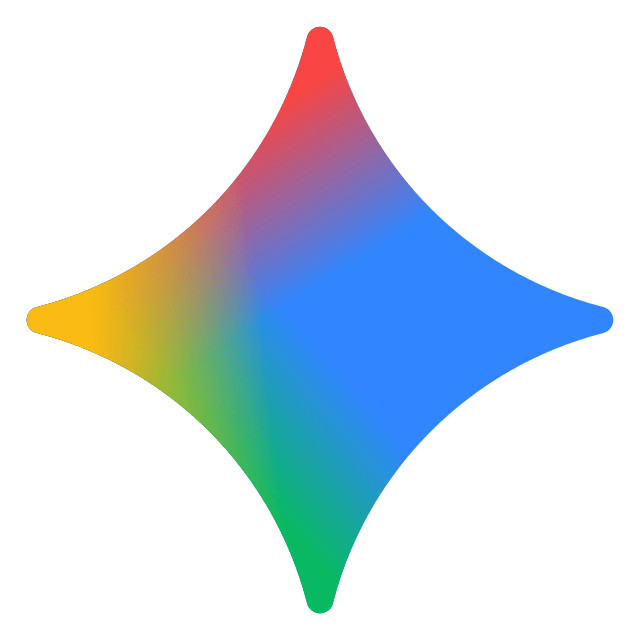}}}
\newcommand{\logoClaude}{\raisebox{-0.2\height}{\includegraphics[height=1.0em]{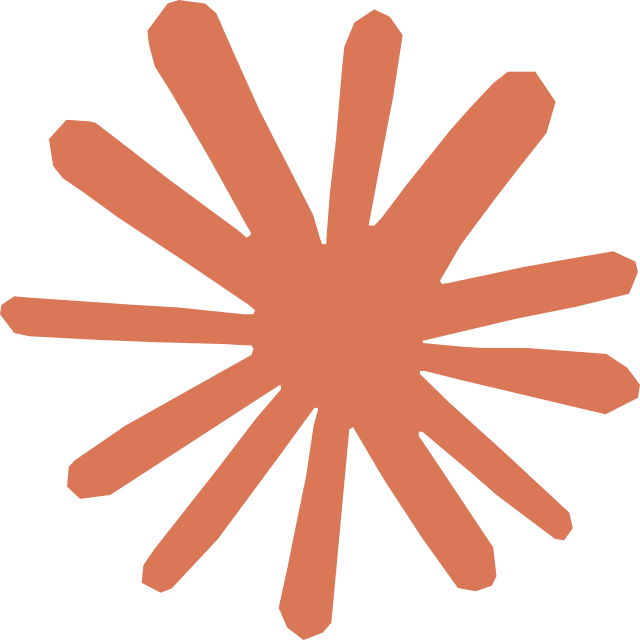}}}
\newcommand{\logoGLM}{\raisebox{-0.2\height}{\includegraphics[height=1.0em]{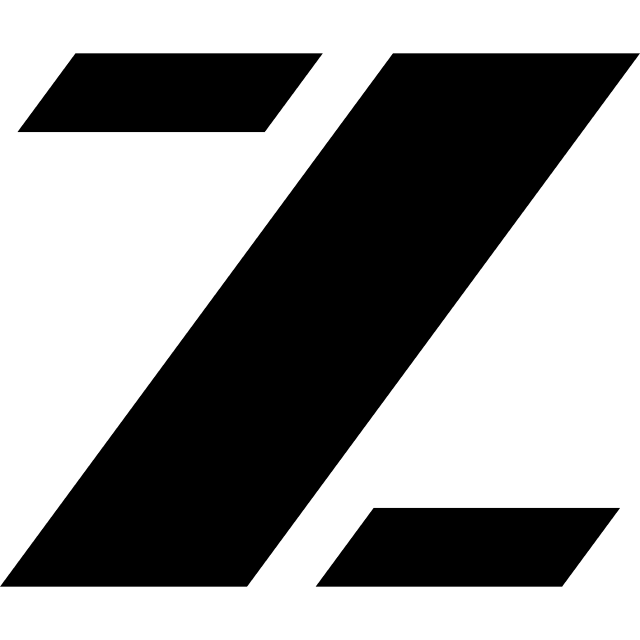}}}
\newcommand{\logoQwen}{\raisebox{-0.2\height}{\includegraphics[height=1.0em]{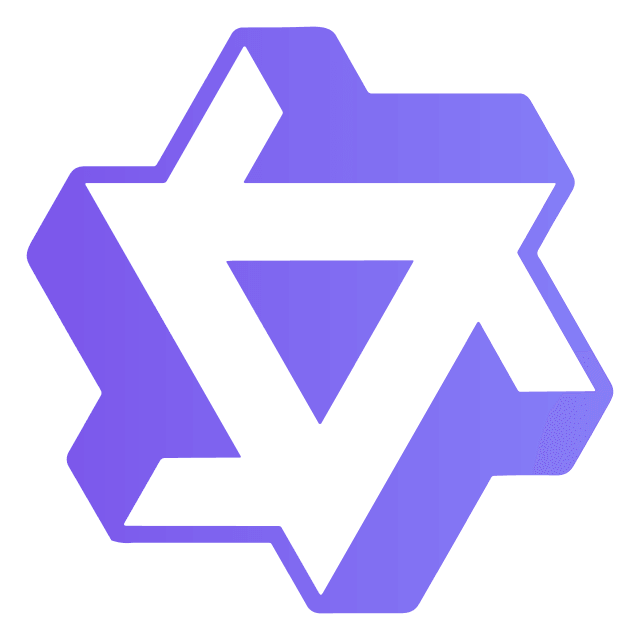}}}
\newcommand{\logoSeed}{\raisebox{-0.2\height}{\includegraphics[height=1.0em]{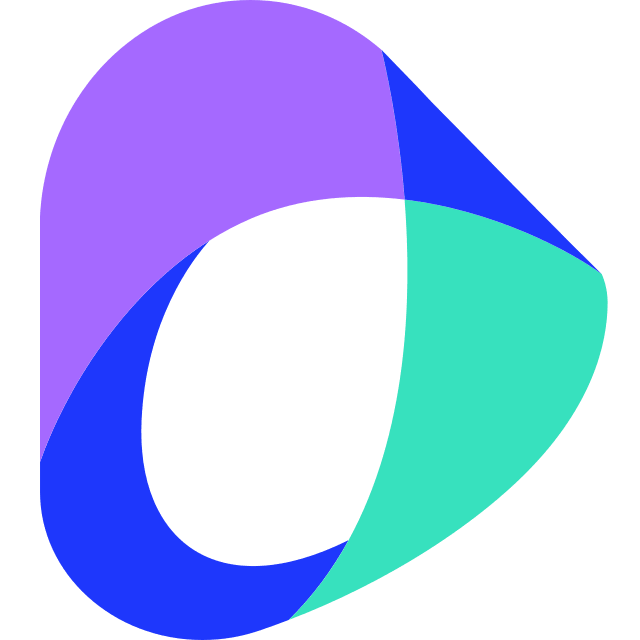}}}
\newcommand{\logoKimi}{\raisebox{-0.2\height}{\includegraphics[height=1.0em]{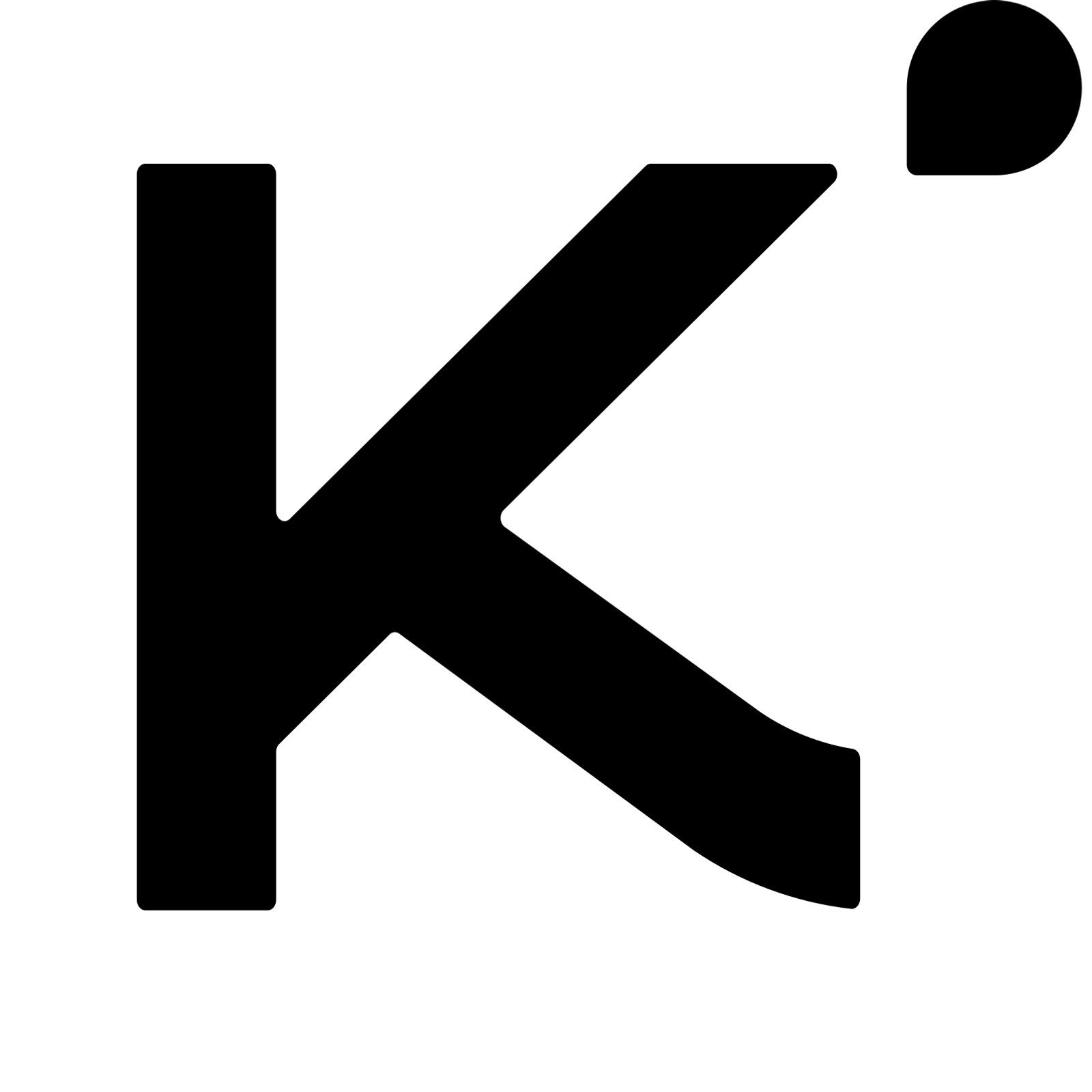}}}
\newcommand{\logoMiniMax}{\raisebox{-0.2\height}{\includegraphics[height=1.0em]{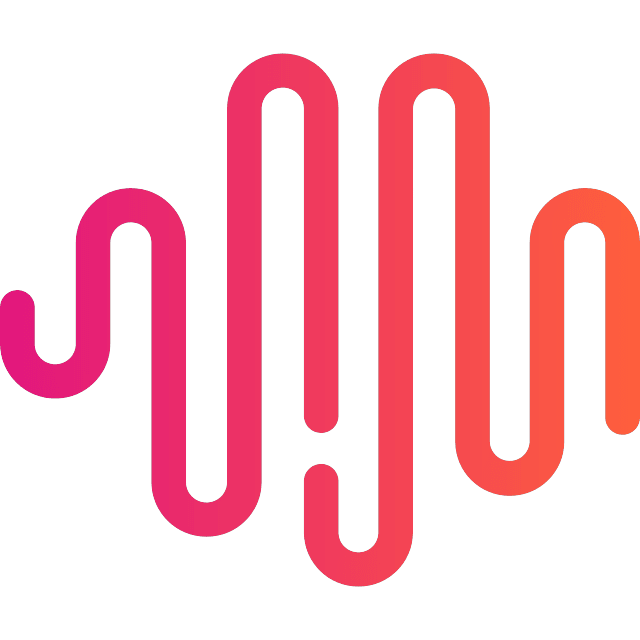}}}
\newcommand{\logoMeta}{\raisebox{-0.2\height}{\includegraphics[height=1.0em]{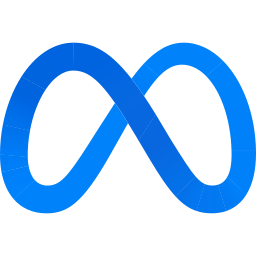}}}

\definecolor{BadgeOpenBg}{RGB}{219,234,254}
\definecolor{BadgeOpenFg}{RGB}{29,78,216}
\definecolor{BadgeOpenBd}{RGB}{147,197,253}
\definecolor{BadgeLowBg}{RGB}{254,226,226}
\definecolor{BadgeLowFg}{RGB}{185,28,28}
\definecolor{BadgeLowBd}{RGB}{252,165,165}
\definecolor{BadgeMedBg}{RGB}{254,243,199}
\definecolor{BadgeMedFg}{RGB}{180,83,9}
\definecolor{BadgeMedBd}{RGB}{252,211,77}
\definecolor{BadgeHighBg}{RGB}{220,252,231}
\definecolor{BadgeHighFg}{RGB}{21,128,61}
\definecolor{BadgeHighBd}{RGB}{134,239,172}
\definecolor{BadgeXHighBg}{RGB}{187,247,208}
\definecolor{BadgeXHighFg}{RGB}{22,101,52}
\definecolor{BadgeXHighBd}{RGB}{74,222,128}
\newcommand{\pillbadge}[4]{%
  \,\ensuremath{\vcenter{\hbox{%
    \tikz{\node[
        fill=#1, text=#2, draw=#3, line width=0.35pt,
        rounded corners=2.4pt,
        inner xsep=3.0pt, inner ysep=1.15pt,
        font=\fontsize{5.5}{6.5}\selectfont\sffamily\bfseries
      ] {#4};}%
  }}}%
}

\newcommand{\badgeLow}{\pillbadge{BadgeLowBg}{BadgeLowFg}{BadgeLowBd}{low}}
\newcommand{\badgeMed}{\pillbadge{BadgeMedBg}{BadgeMedFg}{BadgeMedBd}{medium}}
\newcommand{\badgeHigh}{\pillbadge{BadgeHighBg}{BadgeHighFg}{BadgeHighBd}{high}}
\newcommand{\badgeXHigh}{\pillbadge{BadgeXHighBg}{BadgeXHighFg}{BadgeXHighBd}{xhigh}}

\definecolor{FindingFrame}{RGB}{20,110,110}
\definecolor{FindingFill}{RGB}{245,248,248}
\newcounter{finding}
\renewcommand{\thefinding}{\arabic{finding}}
\newcommand{\finding}[1]{%
  \par\vspace{3pt}\noindent
  \refstepcounter{finding}%
  {\setlength{\fboxsep}{3.5pt}%
   \setlength{\fboxrule}{0.6pt}%
   \fcolorbox{FindingFrame}{FindingFill}{%
     \parbox{\dimexpr\linewidth-2\fboxsep-2\fboxrule\relax}{%
       \textbf{Finding~\thefinding.}~#1}}}%
  \par\vspace{3pt}
}
\definecolor{PyBg}{RGB}{236,245,251}
\definecolor{PyAccent}{RGB}{20,110,110}
\definecolor{PyText}{RGB}{36,48,58}
\definecolor{PyKeyword}{RGB}{0,102,178}
\definecolor{PyString}{RGB}{0,140,100}
\definecolor{PyComment}{RGB}{88,112,122}
\definecolor{PyBuiltin}{RGB}{198,86,20}
\definecolor{PySpecial}{RGB}{124,58,200}
\lstdefinestyle{bvbpython}{
  language={[3]Python},
  basicstyle=\ttfamily\footnotesize\color{PyText},
  classoffset=0,
  keywordstyle=\bfseries\color{PyKeyword},
  classoffset=1,
  keywordstyle=\bfseries\color{PyBuiltin},
  classoffset=0,
  commentstyle=\itshape\color{PyComment},
  stringstyle=\color{PyString},
  literate=
    {Pass}{{{\bfseries\color{PySpecial}Pass}}}{4}
    {Fail}{{{\bfseries\color{PySpecial}Fail}}}{4}
    {Union}{{{\bfseries\color{PySpecial}Union}}}{5}
    {None}{{{\bfseries\color{PySpecial}None}}}{4},
  showstringspaces=false,
  breaklines=true,
  breakindent=1.2em,
  columns=fullflexible,
  keepspaces=true,
  frame=leftline,
  framerule=2.6pt,
  framesep=5pt,
  rulecolor=\color{PyAccent},
  backgroundcolor=\color{PyBg},
  xleftmargin=0.9em,
  xrightmargin=0.55em,
  aboveskip=0.7em,
  belowskip=0.7em,
}
\newcommand{\bvbpageurl}{https://yoloytang.me/BVB/}
\newcommand{\bvbcodeurl}{https://github.com/yunlong10/BVB}

\renewenvironment{abstract}
  {\vskip.03in\noindent\textbf{Abstract:}\hspace{0.35em}\ignorespaces}
  {\par}

\makeatletter
\def\@maketitle{\vbox{\hsize\textwidth
\vbox to 0pt{\vskip 30pt\hbox to \hsize{\hfill
  \includegraphics[height=1.95cm]{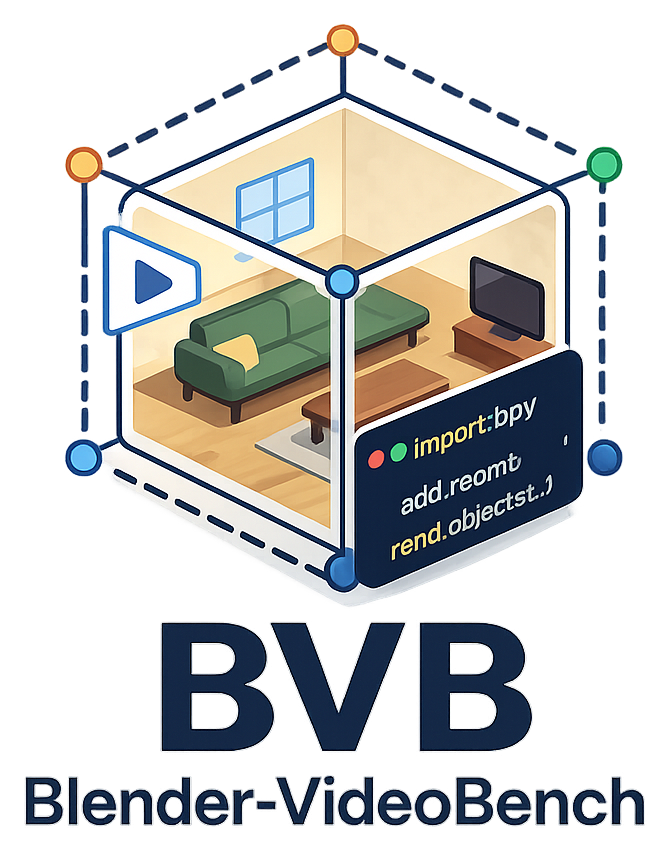}\hspace{-0.14cm}}\vss}%
{\LARGE\bfseries \@title\par}
\setlength{\tabcolsep}{0pt}%
\def\And{\end{tabular}\hfil\linebreak[0]\hfil
        \begin{tabular}[t]{l}\bf\rule{\z@}{24pt}\ignorespaces}%
\def\AND{\end{tabular}\hfil\linebreak[4]\hfil
        \begin{tabular}[t]{l}\bf\rule{\z@}{24pt}\ignorespaces}%
\begin{tabular}[t]{l}\bf\rule{\z@}{24pt}\@author\end{tabular}%
\vskip 0.3in minus 0.1in}}
\makeatother

\makeatletter
\def\section{\@startsection {section}{1}{\z@}{-2.0ex plus
    -0.5ex minus -.2ex}{1.5ex plus 0.3ex
minus0.2ex}{\large\bfseries\raggedright}}
\def\subsection{\@startsection{subsection}{2}{\z@}{-1.8ex plus
-0.5ex minus -.2ex}{0.8ex plus .2ex}{\normalsize\bfseries\raggedright}}
\def\subsubsection{\@startsection{subsubsection}{3}{\z@}{-1.5ex
plus      -0.5ex minus -.2ex}{0.5ex plus
.2ex}{\normalsize\bfseries\raggedright}}
\def\subparagraph{\@startsection{subparagraph}{5}{\z@}{1.5ex plus
  0.5ex minus .2ex}{-1em}{\normalsize\bfseries}}
\makeatother

\definecolor{FrontFill}{HTML}{F4F6F9}
\definecolor{FrontLine}{HTML}{DCE3EB}
\newsavebox{\bvbfrontbox}
\newlength{\bvbframepad}
\makeatletter
\newcommand{\bvbframedfrontmatter}[1]{%
  \begingroup
  \setlength{\textwidth}{\dimexpr\textwidth-2\bvbframepad\relax}%
  \begin{lrbox}{\bvbfrontbox}%
  \begin{minipage}{\textwidth}%
    \@maketitle
    \vskip-0.26in
    #1%
  \end{minipage}%
  \end{lrbox}%
  \noindent\begin{tikzpicture}
    \node[fill=FrontFill, draw=FrontLine, line width=0.7pt,
          rounded corners=7pt, inner sep=\bvbframepad, outer sep=0pt]
         {\usebox{\bvbfrontbox}};
  \end{tikzpicture}%
  \endgroup
}
\makeatother

\definecolor{LinkAccent}{HTML}{7C3AED}
\newcommand{\bvblink}[2]{\href{#1}{\textcolor{LinkAccent}{#2}}}
\newcommand{\bvbsep}{\textcolor{LinkAccent}{\ }\textcolor{LinkAccent}{|}\textcolor{LinkAccent}{\ }}

\title{BVB: Benchmarking Agentic Video Understanding via Programmatic Reconstruction in Blender}

\definecolor{AuthorMarkRed}{HTML}{C62828}
\newcommand{\URmark}{\textsuperscript{\textcolor{AuthorMarkRed}{\ensuremath{\diamondsuit}}}}
\newcommand{\Sonymark}{\textsuperscript{\ensuremath{\clubsuit}}}
\newcommand{\CMUmark}{\textsuperscript{\textcolor{AuthorMarkRed}{\ensuremath{\heartsuit}}}}
\newcommand{\UWmark}{\textsuperscript{\ensuremath{\spadesuit}}}

\author{%
  {\bfseries
    Yolo Y. Tang\URmark \quad
    Daiki Shimada\Sonymark \quad
    Jiayue Meng\URmark \quad
    Jing Bi\URmark \quad
    Pinxin Liu\URmark} \\
  {\bfseries
    Yicheng Wang\CMUmark \quad
    Yunzhong Xiao\CMUmark \quad
    Zhangyun Tan\URmark \quad
    Zeliang Zhang\URmark} \\
  {\bfseries
    Chao Huang\URmark \quad
    Susan Liang\URmark \quad
    Qianxiang Shen\UWmark \quad
    Luchuan Song\URmark} \\
  {\bfseries
    Ali Vosoughi\URmark \quad
    Mingqian Feng\URmark \quad
    Melika Filvantorkaman\URmark \quad
    Chenliang Xu\URmark} \\[2pt]
  \scalebox{0.76}{\normalfont
    \URmark\,University of Rochester \quad
    \Sonymark\,Sony Group Corporation \quad
    \CMUmark\,Carnegie Mellon University \quad
    \UWmark\,University of Washington}
}

\hypersetup{
  pdftitle={BVB: Benchmarking Agentic Video Understanding via Programmatic Reconstruction in Blender},
  pdfauthor={Yolo Y. Tang, Daiki Shimada, Jiayue Meng, Jing Bi, Pinxin Liu, Yicheng Wang, Yunzhong Xiao, Zhangyun Tan, Zeliang Zhang, Chao Huang, Susan Liang, Qianxiang Shen, Luchuan Song, Ali Vosoughi, Mingqian Feng, Melika Filvantorkaman, Chenliang Xu}
}

\begin{document}

\bvbframedfrontmatter{%
  \begingroup
  \hypersetup{hidelinks}
  \centerline{{\ttfamily\small Link:\quad
     \bvblink{\bvbpageurl}{Project Page}\bvbsep
     \bvblink{\bvbcodeurl}{GitHub}}}
  \endgroup
  \vskip 2pt
  \begin{abstract}
  Multimodal agents can create complex videos in software such as
Blender by writing code instead of using diffusion models.
Yet video understanding benchmarks still evaluate models mainly through
question answering.
If an agent truly understands a video, it can reconstruct it programmatically.
We introduce \textbf{BVB}, Blender-VideoBench, a benchmark that tests this
ability by asking agents to reconstruct real-world videos as animated Blender
scenes.
To ensure fair comparison, each agent programs the reconstruction through
a lightweight harness, \textbf{Mini-BVB}, in an identical sandbox under a shared
cost limit.
The benchmark renders each reconstruction from its animated camera and
evaluates it on two axes:
(1)~\textbf{Dual VQA} measures how many spatiotemporal facts
the reconstruction preserves.
(2)~\textbf{Latent Similarity} measures how closely the
reconstruction matches the source video perceptually.
Our overall score, a square-root mean, favors balanced performance.
We evaluate 51 configurations from 10 model families and analyze semantic
retention, perceptual similarity, reasoning effort, and cost.
The best model reaches
88.6 Latent Similarity but retains only 53.7\% of the spatiotemporal
facts from the source video.
Additional reasoning improves perceptual similarity but does not close this gap.
In a blind study with 15 raters and five configurations, Latent Similarity
correlates strongly with human preference.
These results show that programmatic reconstruction is a viable test of
agentic video understanding, and that semantic retention remains the
main challenge.

  \end{abstract}
}
\par\vspace{24pt}
\fancyhead{}
\lhead{\scriptsize\itshape BVB: Benchmarking Agentic Video Understanding via Programmatic Reconstruction in Blender}
\begin{figure}[H]
  \vspace{-2.1em}
  \centering
  \includegraphics[width=\linewidth]{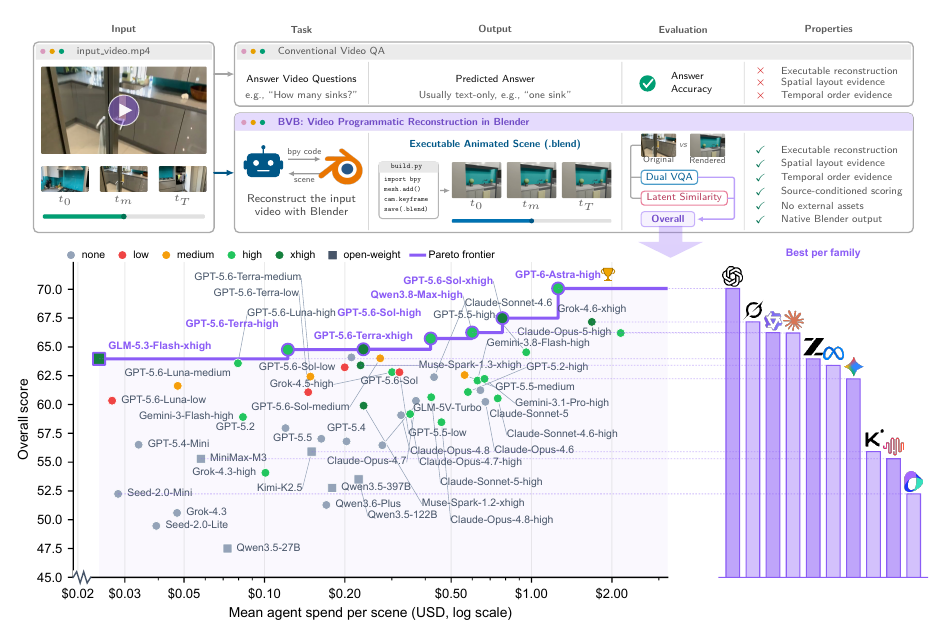}\\[0pt]
  \caption{BVB overview and \SqrtMean{} cost frontier across 51
  configurations.}
  \label{fig:teaser}
  \label{fig:cost_frontier_sqrt}
\end{figure}

\section{Introduction}
\label{sec:intro}
Video understanding is usually measured by question
answering~\citep{tang2025videollmsurvey,jang2017tgifqa,lei2018tvqa,yu2019activitynetqa,xiao2021nextqa,patraucean2023perceptiontest,li2024mvbench,wu2024longvideobench,fu2025videomme,tang2025vidcomposition}, but a correct answer can
come from answer priors or a single frame~\citep{lei2023singleframe}, so it
does not show that the model tracked the scene over time. \looseness=-1
\emph{If an agent truly understands a video, it can reconstruct it programmatically.}
A reconstruction must approximate the source in object placement,
camera trajectory, and the order in which objects appear.
These details cannot be guessed from a single frame or answer prior.
The agent needs to work from video frames alone, without depth maps, segmentation,
or 3D ground truth, so it must infer the full scene from pure 2D observation.
To rebuild a video, an agent must combine spatial, temporal, and compositional
understanding with reasoning and coding.
Existing video benchmarks test these abilities separately.

This test has recently become possible.
Multimodal agents can now construct visual content by writing code instead of using
diffusion models. Recent systems build animated Blender scenes through agent-driven
code~\citep{openai2026astra,ricouard2026astra,yin2026viga,he2025kubrick,blendermcp2025},
suggesting that these agents already have some critical spatiotemporal understanding capabilities.
However, existing results come from selected scenes, often with repeated human
guidance and external asset libraries, so they do not show how reliably an
agent can handle new scenes, how performance changes across model families,
or how well the resulting scene matches the source in layout and dynamics.
A rigorous benchmark should evaluate this ability at scale through holistic reconstruction of real-world videos under a shared protocol without external assets.

To enable such controlled evaluation, we introduce \textbf{BVB}, Blender-VideoBench, a benchmark asking multimodal agents to reconstruct real-world videos as animated Blender scenes, as shown in Figure~\ref{fig:teaser}.
To ensure that agents are required to understand actual scenes, and to
keep reconstruction complexity manageable while preserving rich
spatiotemporal structure, we construct the benchmark with the egocentric real-world indoor videos and
question-answer pairs from VSI-Bench~\citep{yang2025thinkinginspace}.
Each agent interacts through a lightweight harness (\textbf{Mini-BVB}) that offers
two actions, inspecting video frames and executing code in a Blender sandbox,
under a shared cost limit.
External asset libraries are disallowed, so the agent must construct the scene
from primitives, animate its camera along the source trajectory, and save
the result as an editable Blender file instead of a generated image or a
pre-rendered video.

A faithful reconstruction must preserve the source video's layout and dynamics,
so BVB evaluates each reconstruction along two axes:
(1)~\DualVQA{} (\abDV{}) measures how many spatiotemporal facts the reconstruction
preserves. It asks a VLM judge the same spatial and temporal questions on the
source and reconstruction, from object counts and distances to route plans and
appearance order, and scores retention only on questions the judge answers
correctly on the source.
(2)~\LatentSim{} (\abLS{}) measures how closely the reconstruction matches the
source video perceptually, comparing the two videos with frozen V-JEPA~2.1
representations~\citep{bardes2024vjepa,assran2025vjepa2,murlabadia2026vjepa21}.
We report both axes and rank configurations by their square-root mean
(\SqrtMean{}), which favors balanced performance across the two axes.

We evaluate 51 configurations across 10 proprietary and open-weight model
families and analyze semantic retention, perceptual similarity, reasoning
effort, and cost.
GPT-6-Astra-\texttt{high} leads at 70.07 \SqrtMean{} and reaches 88.6 \abLS{},
but it retains only 53.7\% of the spatiotemporal facts from the source video.
Current agents therefore build reconstructions that look right but get many
facts wrong.
Additional reasoning improves perceptual similarity but does not close this gap
in semantic retention.
To validate the automatic evaluation, we conduct a blind study with 15 human raters,
whose mean ranking of five configurations matches the \SqrtMean{} order exactly.
These results show that programmatic reconstruction is already a viable test of
video understanding, but that the best models still miss nearly half the
spatiotemporal facts from the source video.

In short, our contributions are threefold:
\begin{itemize}
    \item We introduce \textbf{BVB}, a benchmark that tests agentic video
    understanding by asking multimodal agents to reconstruct real-world indoor
    videos as animated Blender scenes under a standardized, cost-controlled, asset-free setup.
    \item We design a two-axis evaluation that favors balanced performance across
    semantic retention and perceptual similarity, and validate it against human
    blind rankings.
    \item We evaluate 51 configurations across 10 model families, including
    GPT-6 Astra, and find that current agents produce visually plausible but
    semantically incomplete reconstructions. These results identify which video
    understanding abilities remain unsolved, highlighting open challenges for
    the community.
\end{itemize}

\begin{figure}[H]
  \centering
  \includegraphics[width=\linewidth]{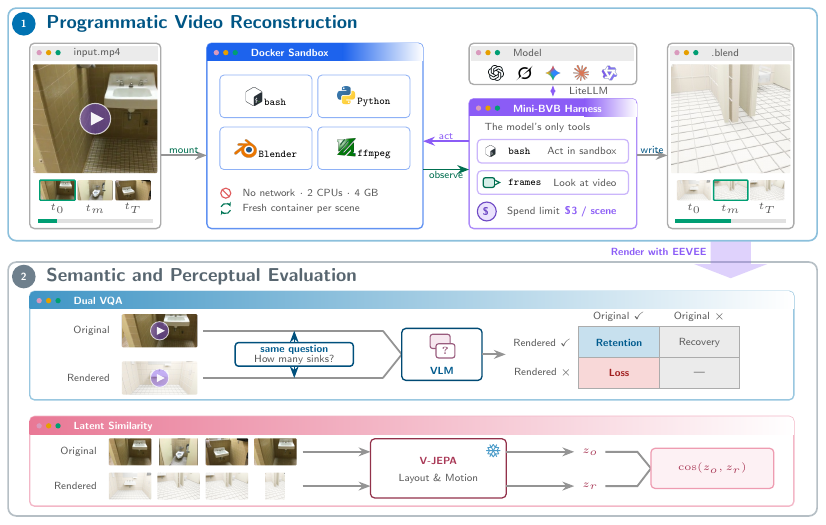}
  \caption{\textbf{BVB agent runs and evaluation.}
  A model alternates between viewing source frames and running Python code in a Docker sandbox
  under a cost limit, then saves an animated Blender scene.
  \DualVQA{} measures how well the rendered video retains answers the judge gets
  right on the source, and \LatentSim{} compares layout and motion using frozen V-JEPA
  features.}
  \label{fig:pipeline}
\end{figure}

\section{BVB: Blender-VideoBench}
\label{sec:benchmark}

\subsection{Benchmark Construction}
\label{subsec:benchmark_construction}

BVB is built on the real indoor egocentric videos of
VSI-Bench~\citep{yang2025thinkinginspace}, drawn from
ARKitScenes~\citep{baruch2021arkitscenes}, ScanNet~\citep{dai2017scannet}, and
ScanNet++~\citep{yeshwanth2023scannetpp}. We use its 288 scenes and their
5,130 QA pairs, so every reconstruction is scored against the same set of questions.
In Stage~1 (Figure~\ref{fig:pipeline}), each agent reconstructs the video in a Docker
sandbox with Blender through a lightweight harness called \textbf{Mini-BVB},
which exposes exactly two actions. \texttt{bash} runs sandbox commands including Python code for Blender, and
\texttt{frames} requests video frames by timestamp or uniform count.
In this agentic setting, the model itself decides how much of the video
to look at and how long to work, with no step or frame budget. The only limit is a
per-scene spend cap.
External asset libraries are disallowed, so geometry must be built from primitives
and basic Blender operations.
A run is valid only if the agent observed the source video and produced a
renderable Blender file. Stage~2 scores this file and never re-enters the loop.
If an agent fails to produce a renderable file, that scene scores zero on every
metric, because excluding failed scenes would let a model raise its average by
skipping hard ones.
Appendix~\ref{app:harness} gives the loop in full and reproduces the system
prompt, which is identical for every configuration, so every model faces the same
harness, sandbox, prompt, and cost ceiling.

\subsection{Evaluation Metrics}
\label{subsec:evaluation_metrics}

BVB evaluates reconstructions on two complementary axes (Figure~\ref{fig:pipeline}).
\textbf{\DualVQA{}} measures how much semantic content the reconstruction retains.
\textbf{\LatentSim{}} measures how closely the reconstruction matches the source in
overall visual appearance.
We combine both axes into \SqrtMean{}.

\paragraph{\DualVQA{} (\abDV{}).}
A faithful reconstruction should retain the spatial and temporal facts of the source.
A VLM judge answers the VSI-Bench questions on
uniformly sampled frames of both the source and rendered video.
The retention rate of \abDV{} is defined as $|C_s \cap C_r|\,/\,|C_s|$, where $C_s$
and $C_r$ are the questions answered correctly on the source and reconstruction.
We refer to the facts tested by $C_s$ as recoverable facts, since the judge can
recover them from the source video.
Conditioning on $C_s$ means the metric captures what the reconstruction keeps,
not the judge's baseline accuracy.

\paragraph{\LatentSim{} (\abLS{}).}
Correct answers to discrete questions do not guarantee that a reconstruction
\emph{looks} like the source, so we add a continuous perceptual axis.
A frozen V-JEPA
encoder~\citep{bardes2024vjepa,assran2025vjepa2,murlabadia2026vjepa21}
maps both clips to latent features. We compare spatial arrangement and
temporal dynamics separately, reporting them as Layout and Motion, and average
them to get \abLS{}.
The encoder is a self-supervised video model, never fine-tuned on BVB, so
the metric reflects general visual agreement rather than benchmark-specific
patterns.
Appendix~\ref{sec:appendix_vjepa} gives the formal definition and pipeline
details.

\begingroup
\setlength{\abovedisplayskip}{4pt}
\setlength{\belowdisplayskip}{4pt}
\setlength{\abovedisplayshortskip}{2pt}
\setlength{\belowdisplayshortskip}{2pt}
\paragraph{\SqrtMean{}.}
The two axes measure different aspects of reconstruction quality. \abDV{} is a
semantic retention rate and \abLS{} is a perceptual similarity score. Under
the arithmetic mean
$\bar{s}_i=1/|\mathcal{A}|\sum_{a\in\mathcal{A}}s_i^a$, a gain on one axis
exactly offsets an equal loss on the other. A 10-point increase in \abLS{} fully
compensates for a 10-point decrease in \abDV{}, which is not the trade-off we
want the aggregate to encode. To favor configurations that are strong on both
axes we instead use the square-root mean:
\begin{equation}
\label{eq:sqrt_mean}
\bar{s}_i =
\left(
\frac{1}{|\mathcal{A}|}
\sum_{a\in\mathcal{A}}\sqrt{s_i^a}
\right)^2,
\qquad
\mathcal{A}=\{\abDV{},\abLS{}\}.
\end{equation}
Both axes are expressed on a 0--100 scale.
Equivalently, the square-root mean is the arithmetic mean corrected by a cross-axis dispersion
penalty,
$\bar{s}_i=1/|\mathcal{A}|\sum_{a\in\mathcal{A}}s_i^a-
\mathrm{Var}_{a\in\mathcal{A}}(\sqrt{s_i^a})$, as
Appendix~\ref{sec:appendix_sqrt_mean} derives.
Equal gains and losses on the two axes therefore no longer cancel exactly,
and more uneven performance receives a larger penalty. Moreover, unlike
the geometric mean, the square-root mean does not become zero when only one axis
is zero. We therefore use it to rank configurations.
\endgroup

\section{Experiments}
\label{sec:experiments}

\subsection{Experimental Setup}
\label{subsec:experimental_setup}
\label{subsec:baselines}

\paragraph{Models.} We evaluate frontier multimodal models as zero-shot reconstruction agents,
without fine-tuning. The suite covers ten families of proprietary API and
open-weight models, namely OpenAI GPT-5.x and GPT-6 Astra~\citep{openai2026astra},
xAI Grok, Anthropic Claude, Google Gemini~\citep{gemini2025gemini25},
Meta Muse Spark~\citep{meta2026musespark13},
Zhipu GLM~\citep{zhipu2025glm45v,zai2026glm53flash}, Alibaba Qwen~\citep{qwen2025qwen25vl},
Moonshot Kimi~\citep{kimi2025k2}, MiniMax~\citep{minimax2025minimax01}, and
ByteDance Seed~\citep{seed2025seed15vl}. For models with adjustable reasoning effort, we test
the available settings among \texttt{none}, \texttt{low}, \texttt{medium}, \texttt{high}, and \texttt{xhigh}.

\paragraph{Implementation details.} All agents run through the same
Mini-BVB harness and Stage~1 sandbox defined in Section~\ref{sec:benchmark}.
Each configuration is evaluated on all 288 scenes using the 5{,}130
spatiotemporal questions.
We use a common \$3 per-scene cost ceiling and the same system
prompt, which shows the agent its source video and requires an executable
Blender program and a final \texttt{result.blend}.
The Docker sandbox comes pre-installed with Bash, Python, Blender~4.2, and FFmpeg.
For \DualVQA{}, the VLM judge is \texttt{gpt-5.4-mini}, answering all
VSI-Bench questions on 16 uniformly sampled frames from both the source and
rendered video.
The judge answers 35.6\% of the questions (1,827 of 5,130) correctly on the
source videos, and these questions form the \abDV{} denominator.
We report retention per task,
covering object counting, absolute and relative distance, sizes, direction,
route planning, and appearance order.
For \LatentSim{}, we render 64 frames from each reconstruction with Blender's
EEVEE renderer along the scene-camera timeline and sample 64 frames from the
source video. The encoder is V-JEPA~2.1
ViT-G~\citep{murlabadia2026vjepa21}.

\subsection{Main Results}
\label{subsec:main_results}

\begin{table}[t]
\centering
\definecolor{ColBest}{RGB}{196,181,253}
\definecolor{ColSecond}{RGB}{221,214,254}
\caption{\textbf{A subset of the BVB results}, with proprietary and
open-weight models listed separately.
\textbf{Rank} is the global ordering over all 51 configurations
(Table~\ref{tab:full_results}).
Colored effort badges encode the reasoning setting (red for \texttt{low}, amber for \texttt{medium},
green for \texttt{high} or \texttt{xhigh}, no badge for \texttt{none}).
{\setlength{\fboxsep}{1pt}\colorbox{ColBest}{\strut Darker}} and
{\setlength{\fboxsep}{1pt}\colorbox{ColSecond}{\strut lighter}} purple mark the best and second-best among shown rows.}
\label{tab:main_results}
\definecolor{ColProp}{RGB}{255,246,220}
\definecolor{ColOpenSec}{RGB}{220,245,230}
\definecolor{RankPurple}{RGB}{124,58,237}
\setlength{\tabcolsep}{1.6pt}
\renewcommand{\arraystretch}{1.05}
\newcommand{\hdrpad}[1]{\rule{0pt}{2.6ex}#1\rule[-1.0ex]{0pt}{0pt}}
\setlength{\aboverulesep}{0pt}
\setlength{\belowrulesep}{0pt}
\resizebox{\linewidth}{!}{%
\begin{tabular}{l|cc|ccccccccc|ccc}
\toprule
\multirow{2}{*}[-3.4ex]{\shortstack[l]{\textbf{Model \&}\\\textbf{Reasoning Effort}}} &
& &
\multicolumn{9}{c|}{\hdrpad{\textbf{\DualVQA{} (\abDV{})} $\uparrow$}} &
\multicolumn{3}{c}{\hdrpad{\textbf{\abLS{}} $\uparrow$}} \\
\cmidrule{4-12}\cmidrule{13-15}
&
\multicolumn{1}{c}{\rotatebox{55}{\small\textbf{Rank}}} &
\multicolumn{1}{c|}{\rotatebox{55}{\small\SqrtMean{}}} &
\multicolumn{1}{c}{\rotatebox{55}{\small Obj.\ Count}} &
\multicolumn{1}{c}{\rotatebox{55}{\small Abs.\ Dist.}} &
\multicolumn{1}{c}{\rotatebox{55}{\small Obj.\ Size}} &
\multicolumn{1}{c}{\rotatebox{55}{\small Room Size}} &
\multicolumn{1}{c}{\rotatebox{55}{\small Rel.\ Dist.}} &
\multicolumn{1}{c}{\rotatebox{55}{\small Rel.\ Dir.}} &
\multicolumn{1}{c}{\rotatebox{55}{\small Route Plan}} &
\multicolumn{1}{c}{\rotatebox{55}{\small Appr.\ Order}} &
\multicolumn{1}{c|}{\rotatebox{55}{\small\textbf{Avg.}}} &
\multicolumn{1}{c}{\rotatebox{55}{\small Layout}} &
\multicolumn{1}{c}{\rotatebox{55}{\small Motion}} &
\multicolumn{1}{c}{\rotatebox{55}{\small\textbf{Avg.}}} \\
\midrule
\multicolumn{4}{l}{\cellcolor{ColProp}\textit{Proprietary Models}} &
\multicolumn{1}{@{}c@{}}{\cellcolor{ColProp!90}} &
\multicolumn{1}{@{}c@{}}{\cellcolor{ColProp!81}} &
\multicolumn{1}{@{}c@{}}{\cellcolor{ColProp!72}} &
\multicolumn{1}{@{}c@{}}{\cellcolor{ColProp!63}} &
\multicolumn{1}{@{}c@{}}{\cellcolor{ColProp!54}} &
\multicolumn{1}{@{}c@{}}{\cellcolor{ColProp!45}} &
\multicolumn{1}{@{}c@{}}{\cellcolor{ColProp!36}} &
\multicolumn{1}{@{}c@{}}{\cellcolor{ColProp!27}} &
\multicolumn{1}{@{}c@{}}{\cellcolor{ColProp!18}} &
\multicolumn{1}{@{}c@{}}{\cellcolor{ColProp!10}} &
\multicolumn{1}{@{}c@{}}{\cellcolor{ColProp!4}} \\
\logoGPT~GPT-6-Astra\badgeHigh & \cellcolor{RankPurple!42}1 & \cellcolor{ColBest}70.07 & \cellcolor{ColBest}48.5 & 33.5 & \cellcolor{ColBest}73.7 & 33.1 & \cellcolor{ColBest}49.0 & 50.0 & 54.8 & \cellcolor{ColBest}36.0 & \cellcolor{ColBest}53.7 & \cellcolor{ColBest}87.3 & \cellcolor{ColBest}90.0 & \cellcolor{ColBest}88.6 \\
\logoGPT~GPT-5.6-Sol\badgeXHigh & \cellcolor{RankPurple!41}2 & \cellcolor{ColSecond}67.49 & 38.0 & 39.9 & \cellcolor{ColSecond}69.0 & \cellcolor{ColBest}53.1 & 43.6 & 50.7 & 47.3 & 16.0 & 51.7 & \cellcolor{ColSecond}84.0 & \cellcolor{ColSecond}86.7 & \cellcolor{ColSecond}85.4 \\
\logoGrok~Grok-4.6\badgeXHigh & \cellcolor{RankPurple!40}3 & 67.17 & 40.5 & 42.2 & 66.2 & \cellcolor{ColSecond}46.9 & 42.7 & 52.5 & 59.1 & 24.0 & 52.1 & 82.9 & 85.5 & 84.2 \\
\logoQwen~Qwen3.8-Max\badgeHigh & \cellcolor{RankPurple!39}4 & 66.24 & \cellcolor{ColSecond}42.5 & 37.6 & 65.8 & 45.4 & 38.6 & \cellcolor{ColBest}56.1 & \cellcolor{ColBest}72.0 & \cellcolor{ColSecond}30.0 & \cellcolor{ColSecond}52.7 & 79.9 & 82.8 & 81.3 \\
\logoClaude~Claude-Opus-5\badgeHigh & \cellcolor{RankPurple!39}5 & 66.21 & 42.0 & \cellcolor{ColBest}46.2 & 66.4 & 43.8 & 44.0 & 52.0 & \cellcolor{ColSecond}65.6 & 16.0 & 52.6 & 79.8 & 82.9 & 81.4 \\
\logoGPT~GPT-5.6-Sol\badgeHigh & \cellcolor{RankPurple!38}6 & 65.73 & 34.5 & 37.6 & 65.0 & 40.8 & 37.8 & 52.2 & 63.4 & 26.0 & 49.8 & 82.5 & 85.3 & 83.9 \\
\logoGPT~GPT-5.6-Terra\badgeHigh & \cellcolor{RankPurple!37}8 & 64.76 & 37.5 & 34.1 & 66.2 & 45.4 & 30.3 & 52.5 & 63.4 & 16.0 & 49.2 & 81.0 & 83.9 & 82.4 \\
\logoGPT~GPT-5.6-Sol & \cellcolor{RankPurple!35}10 & 64.09 & 31.0 & 37.0 & 67.1 & 37.7 & 40.2 & 50.7 & 61.3 & 22.0 & 49.5 & 79.1 & 82.0 & 80.6 \\
\logoGPT~GPT-5.6-Luna\badgeHigh & \cellcolor{RankPurple!33}13 & 63.58 & 28.5 & 36.4 & 65.2 & 34.6 & 37.3 & 54.4 & 54.8 & 10.0 & 48.2 & 79.6 & 82.7 & 81.1 \\
\logoMeta~Muse-Spark-1.3\badgeXHigh & \cellcolor{RankPurple!33}14 & 63.40 & 35.0 & 36.4 & 64.1 & 35.4 & 40.2 & 51.2 & 62.4 & 20.0 & 48.9 & 78.3 & 81.2 & 79.7 \\
\logoGrok~Grok-4.5\badgeHigh & \cellcolor{RankPurple!31}16 & 62.81 & 35.0 & 39.3 & 64.1 & 42.3 & 39.8 & 47.3 & 59.1 & 14.0 & 48.4 & 77.6 & 80.4 & 79.0 \\
\logoGemini~Gemini-3.1-Pro\badgeHigh & \cellcolor{RankPurple!28}21 & 62.23 & 30.0 & \cellcolor{ColSecond}44.5 & 65.2 & 43.1 & 41.1 & \cellcolor{ColSecond}55.1 & 63.4 & 20.0 & 51.1 & 72.9 & 76.1 & 74.5 \\
\logoGemini~Gemini-3.8-Flash\badgeHigh & \cellcolor{RankPurple!27}22 & 62.08 & 41.5 & 35.8 & 65.4 & 37.7 & 42.7 & 44.9 & 51.6 & 18.0 & 48.4 & 76.0 & 78.8 & 77.4 \\
\logoClaude~Claude-Sonnet-5\badgeHigh & \cellcolor{RankPurple!24}27 & 60.63 & 29.0 & 37.0 & 65.4 & 28.5 & 39.8 & 54.7 & 54.8 & 10.0 & 48.3 & 72.7 & 76.1 & 74.4 \\
\logoClaude~Claude-Sonnet-4.6\badgeHigh & \cellcolor{RankPurple!23}28 & 60.53 & 35.5 & 32.4 & 62.8 & 34.6 & 36.9 & 52.9 & 58.1 & 12.0 & 47.7 & 73.4 & 76.5 & 74.9 \\
\logoClaude~Claude-Opus-4.6 & \cellcolor{RankPurple!21}31 & 60.23 & 28.0 & 39.3 & 63.7 & 30.0 & 46.1 & 48.0 & 50.5 & 24.0 & 47.5 & 73.0 & 75.9 & 74.5 \\
\logoGemini~Gemini-3-Flash\badgeHigh & \cellcolor{RankPurple!18}35 & 58.91 & 31.5 & 38.7 & 66.4 & 40.8 & 39.0 & 52.5 & 58.1 & 18.0 & 49.6 & 67.3 & 70.7 & 69.0 \\
\logoClaude~Claude-Opus-4.8\badgeHigh & \cellcolor{RankPurple!18}36 & 58.46 & 29.0 & 33.5 & 60.9 & 30.0 & 41.1 & 51.7 & 54.8 & 16.0 & 46.4 & 69.9 & 73.9 & 71.9 \\
\logoGPT~GPT-5.5 & \cellcolor{RankPurple!16}38 & 57.03 & 21.5 & 24.3 & 59.8 & 30.8 & 34.9 & 46.3 & 59.1 & 14.0 & 42.6 & 71.8 & 75.4 & 73.6 \\
\logoGLM~GLM-5V-Turbo & \cellcolor{RankPurple!14}41 & 56.47 & 25.5 & 42.2 & 61.5 & 36.9 & 32.8 & 52.9 & 59.1 & 12.0 & 46.8 & 65.4 & 68.7 & 67.0 \\
\logoGrok~Grok-4.3\badgeHigh & \cellcolor{RankPurple!12}44 & 54.08 & 22.0 & 36.4 & 57.5 & 32.3 & 41.1 & 51.0 & 52.7 & 12.0 & 44.7 & 62.8 & 65.9 & 64.3 \\
\logoSeed~Seed-2.0-Mini & \cellcolor{RankPurple!10}47 & 52.24 & 17.0 & 31.8 & 61.5 & 20.8 & 31.1 & 52.9 & 61.3 & 4.0 & 43.4 & 59.8 & 64.0 & 61.9 \\
\logoQwen~Qwen3.6-Plus & \cellcolor{RankPurple!10}48 & 51.28 & 20.5 & 31.8 & 54.1 & 30.0 & 29.9 & 48.5 & 61.3 & 12.0 & 41.4 & 60.8 & 63.7 & 62.2 \\
\logoSeed~Seed-2.0-Lite & \cellcolor{RankPurple!8}50 & 49.47 & 10.5 & 33.5 & 57.9 & 30.0 & 32.0 & 48.5 & 52.7 & 8.0 & 41.3 & 56.2 & 60.6 & 58.4%
\\
\multicolumn{4}{l}{\cellcolor{ColOpenSec}\textit{Open-weight Models}} &
\multicolumn{1}{@{}c@{}}{\cellcolor{ColOpenSec!90}} &
\multicolumn{1}{@{}c@{}}{\cellcolor{ColOpenSec!81}} &
\multicolumn{1}{@{}c@{}}{\cellcolor{ColOpenSec!72}} &
\multicolumn{1}{@{}c@{}}{\cellcolor{ColOpenSec!63}} &
\multicolumn{1}{@{}c@{}}{\cellcolor{ColOpenSec!54}} &
\multicolumn{1}{@{}c@{}}{\cellcolor{ColOpenSec!45}} &
\multicolumn{1}{@{}c@{}}{\cellcolor{ColOpenSec!36}} &
\multicolumn{1}{@{}c@{}}{\cellcolor{ColOpenSec!27}} &
\multicolumn{1}{@{}c@{}}{\cellcolor{ColOpenSec!18}} &
\multicolumn{1}{@{}c@{}}{\cellcolor{ColOpenSec!10}} &
\multicolumn{1}{@{}c@{}}{\cellcolor{ColOpenSec!4}} \\
\logoGLM~GLM-5.3-Flash\badgeXHigh & \cellcolor{RankPurple!34}12 & 63.96 & 35.0 & 38.2 & 64.5 & 46.2 & \cellcolor{ColSecond}46.9 & 51.7 & 54.8 & 28.0 & 50.8 & 77.0 & 80.3 & 78.7 \\
\logoKimi~Kimi-K2.5 & \cellcolor{RankPurple!14}42 & 55.91 & 24.0 & 32.9 & 59.2 & 33.8 & 36.9 & 47.5 & 58.1 & 6.0 & 44.0 & 67.7 & 70.8 & 69.2 \\
\logoMiniMax~MiniMax-M3 & \cellcolor{RankPurple!13}43 & 55.29 & 20.0 & 35.8 & 61.5 & 30.8 & 40.2 & 49.5 & 62.4 & 14.0 & 45.6 & 64.3 & 67.6 & 65.9 \\
\logoQwen~Qwen3.5-122B & \cellcolor{RankPurple!12}45 & 53.50 & 16.5 & 32.9 & 63.9 & 25.4 & 29.0 & 52.9 & 57.0 & 8.0 & 44.1 & 62.0 & 65.6 & 63.8 \\
\logoQwen~Qwen3.5-397B & \cellcolor{RankPurple!11}46 & 52.75 & 16.0 & 32.9 & 60.9 & 29.2 & 38.2 & 45.1 & 57.0 & 10.0 & 43.0 & 61.9 & 65.2 & 63.5 \\
\logoQwen~Qwen3.5-27B & \cellcolor{RankPurple!8}51 & 47.50 & 13.5 & 27.2 & 54.5 & 26.2 & 30.7 & 45.3 & 49.5 & 4.0 & 38.6 & 55.8 & 58.9 & 57.3%
\\
\bottomrule
\end{tabular}%
}
\end{table}

\paragraph{Overview.}
Table~\ref{tab:main_results} shows a subset of the results.
Appendix~\ref{app:full_results} lists all 51 configurations.
Across all 51 configurations, \SqrtMean{} spans 47.50--70.07, \abDV{} spans
38.6--53.7, and \abLS{} spans 57.2--88.6.
Better configurations tend to improve on both axes, but neither axis determines the
other.
Most importantly, the best \abDV{} is only 53.7.
Even the best model therefore loses nearly half of the recoverable facts.

\paragraph{Rank and frontier.}
GPT-6-Astra-\texttt{high} leads at 70.07 \SqrtMean{}, followed by
GPT-5.6-Sol-\texttt{xhigh} at 67.49 and Grok-4.6-\texttt{xhigh} at 67.17.
The top-two \SqrtMean{} gap of 2.58 points is statistically significant
(see Appendix~\ref{app:uncertainty} for bootstrap details), confirming that
BVB separates even the strongest models.
The corresponding difference in \abDV{} is not statistically significant.
GLM-5.3-Flash-\texttt{xhigh} is the strongest open-weight configuration at rank 12
and 63.96 \SqrtMean{}, but the leading proprietary configurations still score
higher on both axes.
The best choice also depends on the budget (Section~\ref{subsec:cost_analysis}).
Section~\ref{sec:analysis} examines semantic failure, human alignment,
and reasoning effort.

\finding{Multimodal agents can already understand video through programmatic
reconstruction, but BVB is not saturated and separates models at every price tier.}

\noindent
\begin{minipage}[t]{0.50\linewidth}
\vspace{0pt}
\textbf{Human blind ranking.}
Fifteen human raters ranked 5 anonymized reconstructions against the source
video on nine scenes each, without knowing which model produced which
reconstruction.
Table~\ref{tab:human_rank} shows that their mean ranking matches the
\SqrtMean{} order exactly (Spearman $\rho{=}1.00$).
When measured per scene and per model, their preference correlates strongly
with \abLS{} (Spearman $\rho{=}0.83$).
Appendix~\ref{app:human_study} gives the full protocol and scene-level
calibration.
\end{minipage}\hfill
\begin{minipage}[t]{0.48\linewidth}
\vspace{0pt}
\captionsetup{hypcap=false}
\captionof{table}{\textbf{Blind ranking matches the BVB \SqrtMean{} order} for all 5
selected configurations.}
\label{tab:human_rank}
\footnotesize
\setlength{\tabcolsep}{4pt}
\renewcommand{\arraystretch}{1.05}
\begin{tabular}{@{}lccc@{}}
\toprule
\shortstack[l]{\textbf{Model \&}\\\textbf{Reasoning Effort}} & \shortstack[l]{\textbf{Human}\\\textbf{rank} $\downarrow$} &
\shortstack[l]{\textbf{Mean}\\\textbf{rank} $\downarrow$} &
\shortstack[l]{\textbf{BVB}\\\textbf{rank} $\downarrow$} \\
\midrule
\logoGPT~GPT-5.6-Sol\badgeXHigh & \textbf{1} & \textbf{1.47} & \textbf{2} \\
\logoGrok~Grok-4.5\badgeHigh & 2 & 2.23 & 16 \\
\logoGemini~Gemini-3.1-Pro\badgeHigh & 3 & 3.03 & 21 \\
\logoClaude~Claude-Opus-4.6 & 4 & 3.70 & 31 \\
\logoQwen~Qwen3.5-397B & 5 & 4.58 & 46 %
\\
\bottomrule
\end{tabular}
\end{minipage}

\section{Analysis}
\label{sec:analysis}

Section~\ref{sec:experiments} ranks configurations by \SqrtMean{}, but
a single score does not show which spatial and temporal facts are lost, whether the
metrics match human judgment, or how reasoning effort and cost affect
the results. We examine each of these questions below.

\subsection{How large is the gap between looking right and being right?}
\label{subsec:saturation}

The best \abLS{} reaches 88.6, yet the best \abDV{} is only 53.7.
In absolute terms, 981 of the 1,827 spatiotemporal questions that the VLM judge
answers correctly on the source video are still answered correctly after
reconstruction.
The remaining 846 questions are lost even by the strongest model.
High perceptual similarity from the frozen V-JEPA encoder therefore does not
guarantee that the reconstruction preserves the spatial and temporal facts of
the source.

\begin{figure}[t]
  \centering
  \begin{minipage}[b]{0.49\linewidth}
    \centering
    \includegraphics[width=\linewidth]{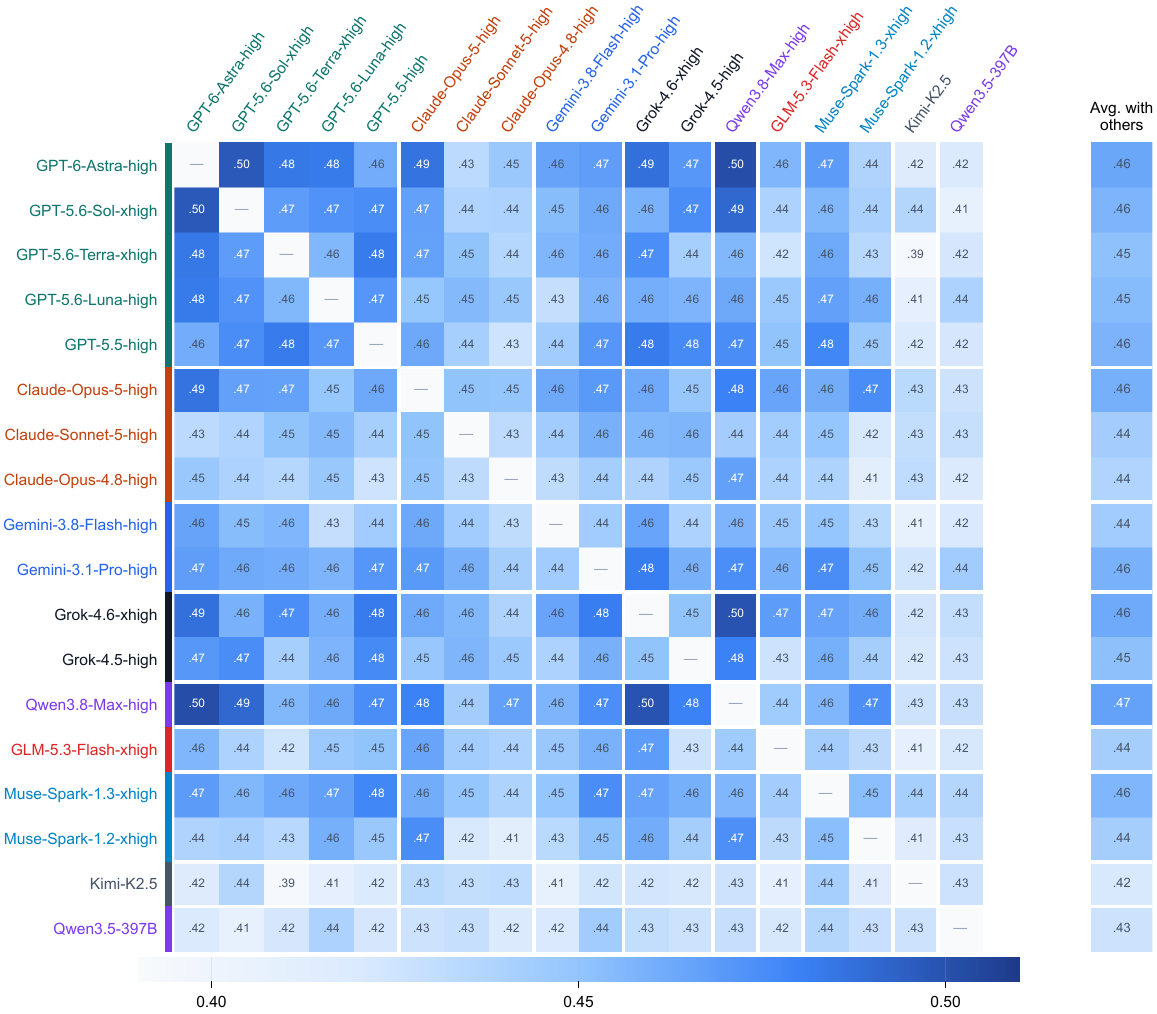}\\[-1pt]
    {\footnotesize (a) Retained-answer overlap}
  \end{minipage}\hfill
  \begin{minipage}[b]{0.49\linewidth}
    \centering
    \includegraphics[width=\linewidth]{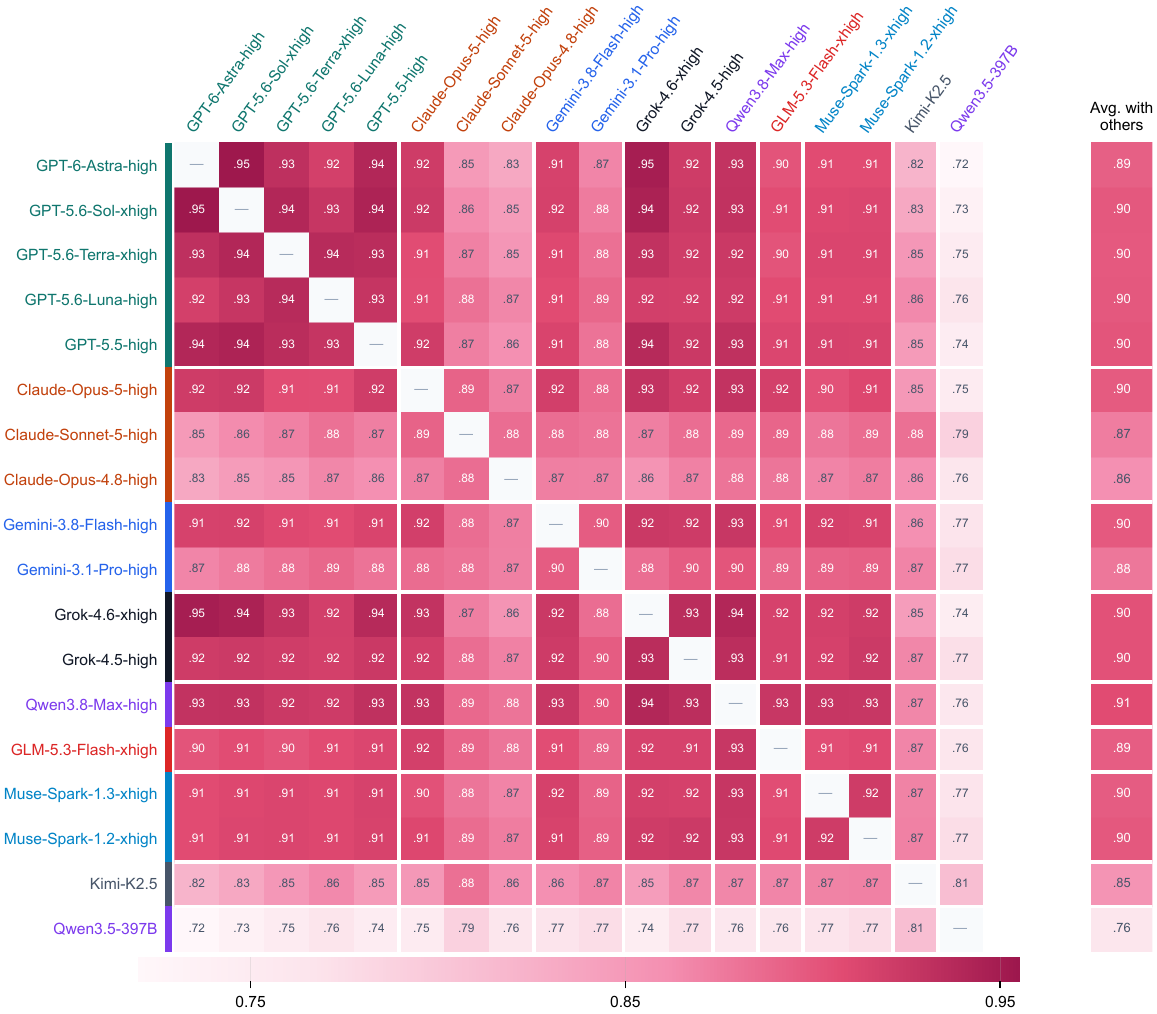}\\[-1pt]
    {\footnotesize (b) V-JEPA similarity}
  \end{minipage}
  \caption{\textbf{Reconstructions converge in appearance but diverge in
  semantic content.}
  (a)~Jaccard overlap of the recoverable facts retained by each pair.
  (b)~V-JEPA similarity between renders.}
  \label{fig:cross_model_similarity}
\end{figure}

Figure~\ref{fig:cross_model_similarity} compares reconstructions across agents.
Across the 153 pairs formed by the 18 configurations, the mean V-JEPA similarity
between rendered videos is 0.88, whereas the mean Jaccard overlap between the
recoverable facts they retain is only 0.45.
This overlap is still well above the 0.33 expected if each agent retained
facts independently at its own rate, and every pair exceeds its expected value.
Agents therefore share a common core of easily retained facts.
Appearance also follows model family more closely than retained facts do.
Same-family pairs average 0.91 in V-JEPA similarity, compared with 0.88 for
cross-family pairs, while their Jaccard overlap rises only from 0.45 to 0.46.
The most similar-looking pair, GPT-6-Astra-\texttt{high} and
GPT-5.6-Sol-\texttt{xhigh}, belongs to one family, whereas the pair sharing the
most retained facts, GPT-6-Astra-\texttt{high} and Qwen3.8-Max-\texttt{high},
spans two.
Stronger agents also converge. Among the 18 configurations, pairs within the top
nine by \SqrtMean{} average 0.93 in V-JEPA similarity and 0.47 in Jaccard
overlap, compared with 0.86 and 0.44 within the bottom nine.
Because most same-family pairs involve strong GPT models, part of the family
effect reflects this convergence.
Visual agreement between agents therefore overstates how much scene content
they share.

\finding{Looking right is not the same as being right.
The best model reaches 88.6 \abLS{} but retains only 53.7\% of the
recoverable facts, so nearly half of them are still lost after reconstruction.}

\subsection{What spatiotemporal information is retained after reconstruction?}
\label{subsec:semantic_retention}

\begin{figure}[t]
  \centering
  \includegraphics[width=\linewidth]{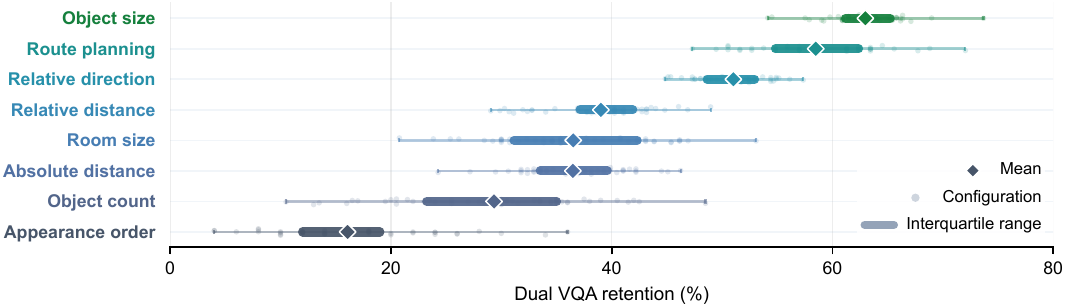}
  \caption{\textbf{Retention varies widely across tasks.} Full range, interquartile
  range, and mean \abDV{} over all 51 configurations for each task.}
  \label{fig:semantic_profiles}
\end{figure}

Figure~\ref{fig:semantic_profiles} summarizes task-level DV distributions
across all configurations.
Across all 51 configurations, object size and route planning have the highest
mean retention at 63.0\% and 58.5\%, while appearance order and object count
are lowest at 16.1\% and 29.3\%.
The benchmark therefore exposes a shared hierarchy of task difficulty.
Tasks that depend on more scene structure have lower retention.
Object size is a property of a single object, and it has the highest
retention.
Object counting requires enumerating every instance in the room, and
appearance order requires tracking the full camera trajectory.
Both tasks lose most of their originally correct answers after
reconstruction.
The spread across models also varies by task.
Object count spans 10.5--48.5\% across configurations, room size spans
20.8--53.1\%, and appearance order spans 4.0--36.0\%.
Relative direction is much more compressed at 44.9--57.4\%.
Table~\ref{tab:main_results} further shows that no configuration dominates
every task.
Thus BVB contains both shared difficulty patterns and task-specific model rankings.

\finding{Task difficulty follows a consistent order across models, yet no single
model is best on every task.
Retention is highest for single-object properties such as size, and lowest
for tasks that require reasoning over the whole scene, such as counting and
appearance order.}

\subsection{Are the metrics complementary and human-aligned?}
\label{subsec:human_alignment}

\begin{figure}[t]
  \centering
  \includegraphics[width=0.95\linewidth]{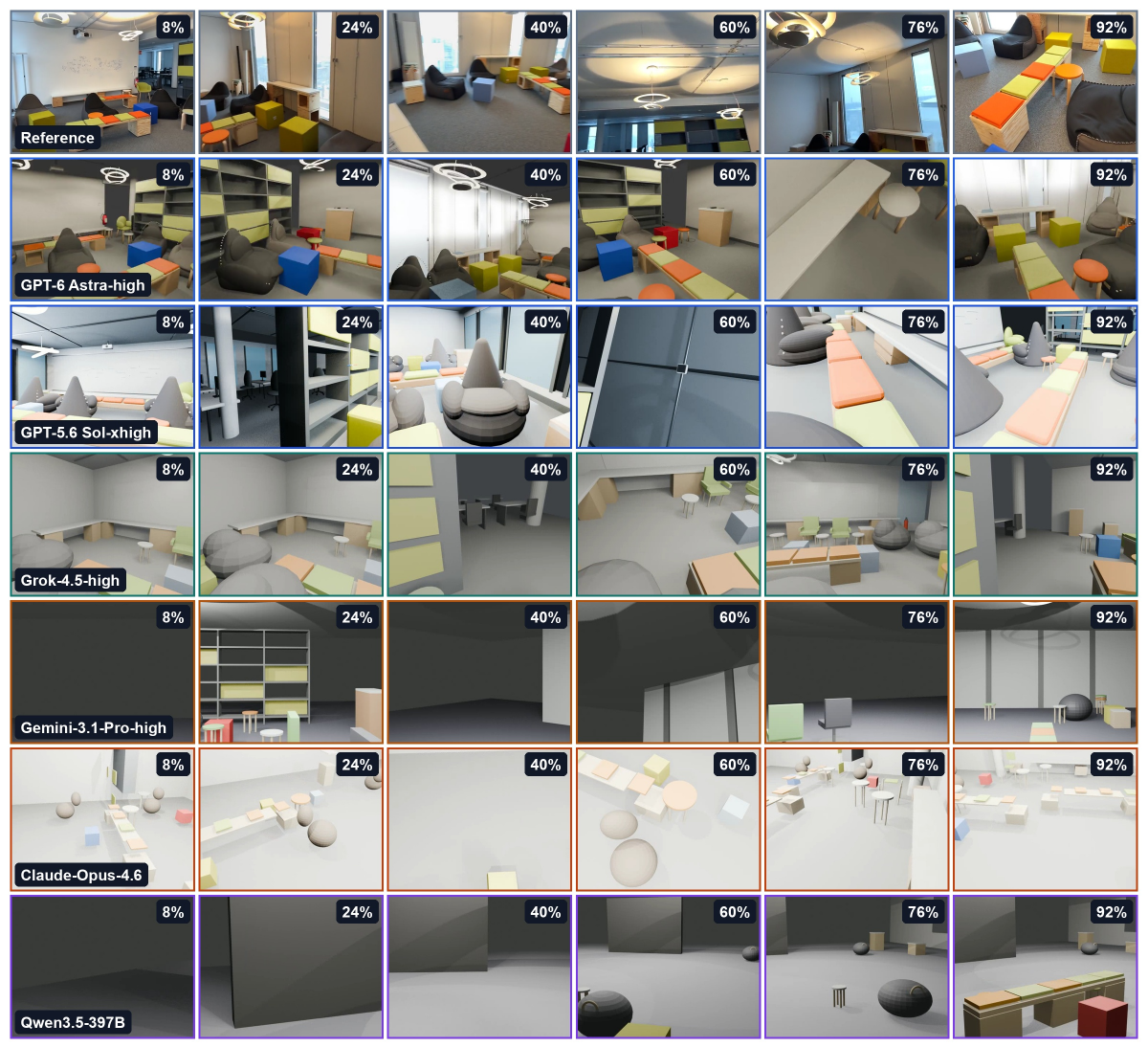}
  \caption{\textbf{Agents reconstruct the same video very differently.}
  Columns are matched video times.}
  \label{fig:qualitative_cross_model}
\end{figure}

\abDV{} and \abLS{} improve together overall, but they rank models
differently. Thus, we report both axes beside \SqrtMean{}.
Figure~\ref{fig:qualitative_cross_model} shows representative examples.
Given the same source video, each model reconstructs different objects,
different layouts, and different portions of the timeline.
The human blind ranking from Section~\ref{subsec:main_results} lets us test
whether the automatic metrics agree with human preferences.
At the scene-model level, \abLS{} correlates with human preference at
Spearman $\rho{=}0.83$
(Appendix Figure~\ref{fig:human_metric_calibration}).
\abDV{} is much weaker at $\rho{=}0.11$.
Human rankings align more strongly with perceptual similarity than with
semantic retention in our study.
At the model level, \SqrtMean{} preserves the complete human ordering of the
five tested configurations (Spearman $\rho{=}1.00$).
The two metrics therefore serve different roles. \abLS{} tracks perceptual
preference, while \abDV{} ensures that visually similar reconstructions do
not receive a high score when they lose factual content.
Appendix~\ref{app:score_diagnostics} examines the questions that the judge
answers incorrectly on the source video.

\finding{\abDV{} and \abLS{} are related but distinct.
Reconstructions reproduce appearance more reliably than factual content.
\abLS{} tracks human preference, while \abDV{} measures semantic retention.}

\subsection{How do reasoning effort and runtime affect scores?}
\label{subsec:effort_runtime}

\begin{figure}[t]
  \centering
  \begin{minipage}[b]{0.61\linewidth}
    \centering
    \includegraphics[width=\linewidth]{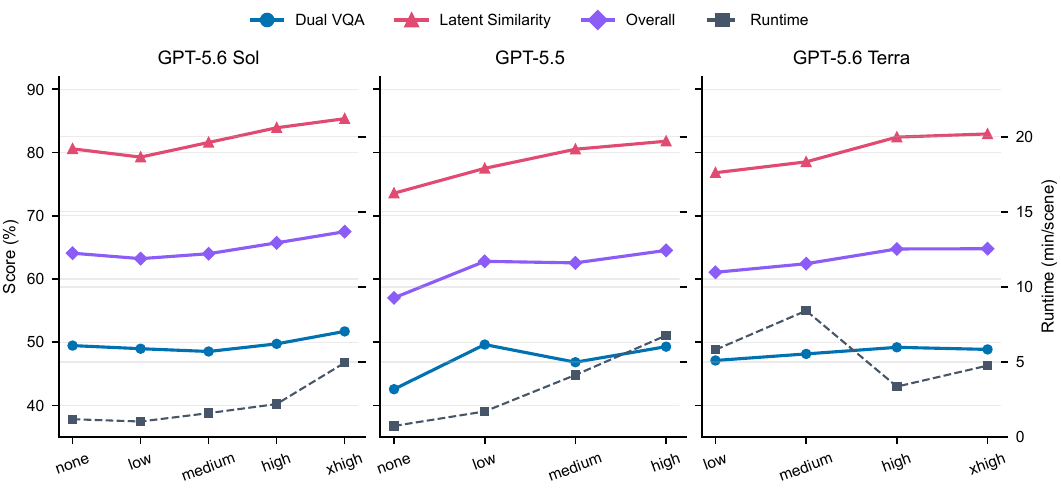}\\[-1pt]
    {\footnotesize (a) Provider effort ladders}
  \end{minipage}\hfill
  \begin{minipage}[b]{0.37\linewidth}
    \centering
    \includegraphics[width=\linewidth]{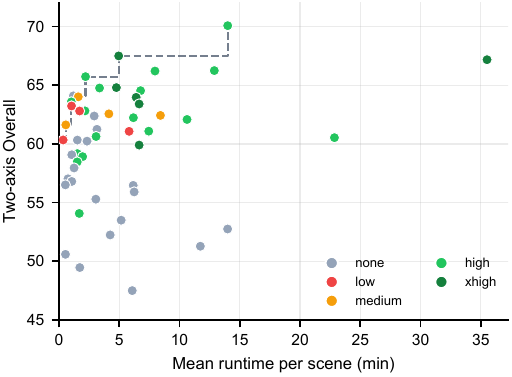}\\[-1pt]
    {\footnotesize (b) Runtime versus \SqrtMean{}}
  \end{minipage}
  \caption{\textbf{Reasoning effort and runtime show different scaling behavior.}
  LS rises more consistently than DV across provider effort ladders, while
  \SqrtMean{} combines the two axes and still reverses at intermediate steps.
  Longer runtime alone does not mean higher \SqrtMean{}.}
  \label{fig:effort_runtime}
\end{figure}

Provider reasoning controls do not change the two scores equally
(Figure~\ref{fig:effort_runtime}a).
Across the GPT-5.6 Sol, GPT-5.5, and GPT-5.6 Terra ladders,
\abLS{} generally rises with effort, whereas \abDV{} stays flat or even
decreases.
A likely explanation is that additional reasoning helps the model refine
geometry, materials, and camera motion, all of which improve visual
similarity, but does not lead the model to verify factual details such as
object counts or spatial relations.
\SqrtMean{} improves from the lowest to the highest available effort in each
family, but the intermediate steps do not always increase.
Runtime behaves similarly (Figure~\ref{fig:effort_runtime}b). Slower configurations
do not consistently score higher, and several fall below the \SqrtMean{} frontier
set by faster ones.
Extra inference time is therefore not a sufficient explanation for score
differences and not a reliable indicator of reconstruction quality.

\finding{Across the tested effort ladders, higher reasoning effort generally
improves \abLS{} more reliably than \abDV{}.
Longer runtime does not mean higher \SqrtMean{}.}

\subsection{What does the frontier cost?}
\label{subsec:cost_analysis}

Mean Stage-1 spend ranges from \$0.024 to \$2.157 per scene, so the most expensive configuration costs roughly
90 times more than the cheapest (Figure~\ref{fig:cost_frontier_sqrt}).
Spending more does not always improve quality.
The top \SqrtMean{} configuration costs \$1.258 per scene, while
GPT-5.6-Sol-\texttt{xhigh} reaches 96\% of that score at \$0.778 and
GLM-5.3-Flash-\texttt{xhigh} reaches 91\% at \$0.024.
The cost difference is driven mainly by model family and reasoning effort
level, not by scene difficulty.
The \abDV{} and \abLS{} frontiers differ across price tiers
(Appendix Figure~\ref{fig:cost_frontier_all}), so the most cost-effective
configurations depend on the evaluation axis.
Evaluating one top-ranked configuration on all 288 scenes costs about \$360,
and most configurations cost much less, so evaluating a new model is affordable.

\finding{Higher cost does not guarantee proportional quality gains.
Configurations at a fraction of the top price still reach over 90\% of the best \SqrtMean{}.}

\section{Related Work}
\label{sec:related_work}

\paragraph{Benchmarking video understanding.}
Most video-understanding benchmarks evaluate models through question answering.
VSI-Bench~\citep{yang2025thinkinginspace} probes visual-spatial intelligence
in real indoor captures and finds that spatial reasoning remains a hard
problem. EgoSchema~\citep{mangalam2023egoschema} targets long-form video
understanding, MMPerspective~\citep{tang2025mmperspective} probes perspective
understanding, and Eagle~\citep{bi2024eagle} targets egocentric video.
Questions in this format can often be answered from static appearance alone,
as single-frame models can match multi-frame
ones~\citep{lei2023singleframe}.
A second family asks agents to act on video rather than answer questions
about it. VideoGUI~\citep{lin2024videogui} evaluates GUI automation from
instructional video,
VideoWebArena~\citep{jang2025videowebarena} evaluates web tasks that need
video evidence, and
ScreenSpot-Pro~\citep{li2025screenspotpro} and
GUIXplore~\citep{sun2025guixplore} test screen grounding across
applications. These benchmarks score task completion in 2D screen
environments.
BVB instead asks the agent to rebuild the video as an editable 3D scene
whose render preserves its layout and dynamics.

\paragraph{Programs as visual representations.}
Writing code is an increasingly common intermediate step in visual reasoning
because it produces a persistent artifact that can be executed and revised.
ViperGPT~\citep{suris2023vipergpt} and VISPROG~\citep{gupta2023visprog}
compose perception modules into generated Python and read answers from
execution traces.
A second line generates programs that rebuild depicted content.
SceneCraft~\citep{hu2024scenecraft} synthesizes 3D scenes from layout descriptions,
VIGA~\citep{yin2026viga} iteratively writes, renders, and revises Blender code,
and Kubrick~\citep{he2025kubrick} coordinates multimodal agents that compose
Blender scripts for synthetic video.
BlenderGym~\citep{gu2025blendergym} benchmarks graphics editing across
placement, lighting, and geometry.
EZBlender~\citep{wang2026ezblender} decomposes editing instructions with a
Plan-and-ReAct agent, and BlenderMCP~\citep{blendermcp2025} exposes Blender
as a language model tool.
Related systems address spatial
variables~\citep{chen2026spatialcode}, simulator
code~\citep{visphyworld2026}, vector
animation~\citep{yang2026omnilottie}, and top-down room
synthesis~\citep{yang2026codeasroom}.
SceneActBench~\citep{zhao2026sceneactbench} decomposes Blender tasks with
hidden geometric ground truth, offering stronger per-task supervision.
BVB instead targets holistic reconstruction from uncalibrated real video,
supplies no assets, and relies on source-conditioned semantic and perceptual
evaluation.

\section{Conclusion}
\label{sec:conclusion}
We propose BVB, which evaluates agentic video
understanding by asking multimodal agents to reconstruct real-world
videos in Blender.
We evaluate 51 configurations across 10 model families, including
GPT-6 Astra, and find that current agents produce visually plausible
but semantically incomplete reconstructions.
Our two-axis evaluation favors balanced performance across
semantic retention and perceptual similarity, and we validate it against
human preferences. Our results also show which video understanding
abilities remain unsolved and where future work is most needed.

\subsubsection*{Acknowledgements}

This work was supported by Sony Group Corporation. We would like to thank
Naofumi Akimoto, Tamaki Kojima, Sayaka Nakamura, and Jerry Jun Yokono for their
insightful discussions.

\bibliography{references}

\appendix
\section{Stage-1 Harness and Agent Prompt}
\label{app:harness}

Section~\ref{subsec:benchmark_construction} describes the Stage-1 loop.
The agent runs on the host and drives a fresh Blender~4.2 Docker container for
each scene.
At every turn it may execute a \texttt{bash} block or request video frames with
a \texttt{frames} block.
The harness sets no turn or frame budget.
The only effort limit is a per-scene dollar budget computed from provider usage.
When cumulative cost reaches the limit the harness terminates the loop
immediately without an extra save turn; a low-budget warning at 25\% remaining
gives the model advance notice to save.
A run finishes normally when the model has consumed at least one frame, saved
\texttt{/workspace/output/result.blend}, and returned \texttt{DONE}.
Evaluation never feeds information back into this loop.

\paragraph{System prompt.}
Every configuration receives the prompt below unchanged.
Only the video and dollar limit vary by scene.

\begin{lstlisting}[language={},basicstyle=\ttfamily\scriptsize\color{PyText},%
  keywordstyle={},commentstyle={},stringstyle={},literate={},%
  breaklines=true,columns=fullflexible,keepspaces=true]
You are a video-understanding agent evaluated on reconstructing an indoor scene
from a video as Blender Python (bpy). You drive a Linux sandbox with Blender
installed.

Interface:
- ```bash block: the harness runs it in the sandbox and returns stdout/stderr.
  Run Blender headless as: blender_run --python /workspace/scratch/build.py
  (wraps `xvfb-run -a blender --background`).
- ```frames block: request video frames to look at. The harness extracts them
  and shows them to you as images on the next turn. Content is either explicit
  timestamps in seconds (e.g. `0, 4.5, 12, 30`) or `count=N` for N evenly-spaced
  frames (optionally with `start=` `end=` seconds). Look at as many frames as
  you need; request more whenever you want.
- When the scene is finished and saved, reply with the single word DONE.

Work in this order - do NOT skip looking at the video:
1. FIRST request frames with a ```frames block, on their own, and WAIT. Do not
   write any bpy or say DONE in the same message as a ```frames request - the
   frames you request are only shown to you on the next turn.
2. Look at the returned frames, then build the scene with ```bash.
3. Request more frames to check details whenever useful, then revise.
4. Only after you have actually seen frames and saved result.blend, say DONE.
A scene built without looking at any frame is a failure.

You decide how much effort to spend: how many frames, how many turns. The raw
video is also at /workspace/video/video.mp4 inside the sandbox.

Scene/output rules:
- The final scene MUST be saved to /workspace/output/result.blend.
- Build ALL geometry from basic primitives only (cube, plane, cylinder, cone,
  uv_sphere, torus) and assemblies of them. Group each object's parts in a
  Collection named after the object.
- No imported models, no sculpting/arbitrary meshes, no geometry nodes,
  particles, physics, or image textures. Materials = a single Principled BSDF
  with numeric values only.
- 1 Blender unit = 1 meter. Keep Unit Scale = 1.0. Use radians for rotations.
- Include at least one camera and one light.
- Reproduce object counts, sizes, positions, and spatial relationships as
  faithfully as you can from the video.

Camera trajectory / temporal reconstruction:
- Reconstruct camera motion as an ANIMATION, not a single static viewpoint.
- Insert keyframes for camera location and rotation_euler over time and set
  frame_start / frame_end to span the motion.
\end{lstlisting}

\section{Full Leaderboard}
\label{app:full_results}

Table~\ref{tab:full_results} reports all 51 configurations.
\SqrtMean{} is the two-axis square-root mean in
Equation~\eqref{eq:sqrt_mean}, and ranks are global over this full set.
The main text retains representative peak and default configurations plus all
open-weight ones.

\begin{table}[p]
\centering
\caption{Full BVB results over 51 configurations.
Dual VQA reports retention on the judge-correct source subset, and Latent
Similarity reports frozen V-JEPA layout and motion agreement.
OK\% is the percentage of scenes that produced a renderable file.
Darker and lighter purple mark the best and second-best values.}
\label{tab:full_results}
\definecolor{ColProp}{RGB}{255,246,220}
\definecolor{ColOpenSec}{RGB}{220,245,230}
\definecolor{RankPurple}{RGB}{124,58,237}
\definecolor{ColBest}{RGB}{196,181,253}
\definecolor{ColSecond}{RGB}{221,214,254}
\setlength{\tabcolsep}{1.5pt}
\renewcommand{\arraystretch}{1.02}
\providecommand{\hdrpad}[1]{\rule{0pt}{2.6ex}#1\rule[-1.0ex]{0pt}{0pt}}
\setlength{\aboverulesep}{0pt}
\setlength{\belowrulesep}{0pt}
\resizebox{\linewidth}{!}{%
\begin{tabular}{l|cc|ccccccccc|ccc|c}
\toprule
\multirow{2}{*}[-3.4ex]{\shortstack[l]{\textbf{Model \&}\\\textbf{Reasoning Effort}}} &
& &
\multicolumn{9}{c|}{\hdrpad{\textbf{\DualVQA{} (\abDV{})} $\uparrow$}} &
\multicolumn{3}{c|}{\hdrpad{\textbf{\abLS{}} $\uparrow$}} & \\
\cmidrule{4-12}\cmidrule{13-15}
&
\multicolumn{1}{c}{\rotatebox{55}{\small\textbf{Rank}}} &
\multicolumn{1}{c|}{\rotatebox{55}{\small\SqrtMean{}}} &
\multicolumn{1}{c}{\rotatebox{55}{\small Obj.\ Count}} &
\multicolumn{1}{c}{\rotatebox{55}{\small Abs.\ Dist.}} &
\multicolumn{1}{c}{\rotatebox{55}{\small Obj.\ Size}} &
\multicolumn{1}{c}{\rotatebox{55}{\small Room Size}} &
\multicolumn{1}{c}{\rotatebox{55}{\small Rel.\ Dist.}} &
\multicolumn{1}{c}{\rotatebox{55}{\small Rel.\ Dir.}} &
\multicolumn{1}{c}{\rotatebox{55}{\small Route Plan}} &
\multicolumn{1}{c}{\rotatebox{55}{\small Appr.\ Order}} &
\multicolumn{1}{c|}{\rotatebox{55}{\small\textbf{Avg.}}} &
\multicolumn{1}{c}{\rotatebox{55}{\small Layout}} &
\multicolumn{1}{c}{\rotatebox{55}{\small Motion}} &
\multicolumn{1}{c|}{\rotatebox{55}{\small\textbf{Avg.}}} &
\multicolumn{1}{c}{\rotatebox{55}{\small\textbf{OK\%}}} \\
\midrule
\multicolumn{4}{l}{\cellcolor{ColProp}\textit{Proprietary Models}} &
\multicolumn{1}{@{}c@{}}{\cellcolor{ColProp!90}} &
\multicolumn{1}{@{}c@{}}{\cellcolor{ColProp!81}} &
\multicolumn{1}{@{}c@{}}{\cellcolor{ColProp!72}} &
\multicolumn{1}{@{}c@{}}{\cellcolor{ColProp!63}} &
\multicolumn{1}{@{}c@{}}{\cellcolor{ColProp!54}} &
\multicolumn{1}{@{}c@{}}{\cellcolor{ColProp!45}} &
\multicolumn{1}{@{}c@{}}{\cellcolor{ColProp!36}} &
\multicolumn{1}{@{}c@{}}{\cellcolor{ColProp!27}} &
\multicolumn{1}{@{}c@{}}{\cellcolor{ColProp!18}} &
\multicolumn{1}{@{}c@{}}{\cellcolor{ColProp!10}} &
\multicolumn{1}{@{}c@{}}{\cellcolor{ColProp!4}} &
\multicolumn{1}{@{}c@{}}{} \\
\logoGPT~GPT-6-Astra\badgeHigh & \cellcolor{RankPurple!42}1 & \cellcolor{ColBest}70.07 & \cellcolor{ColBest}48.5 & 33.5 & \cellcolor{ColBest}73.7 & 33.1 & \cellcolor{ColBest}49.0 & 50.0 & 54.8 & \cellcolor{ColBest}36.0 & \cellcolor{ColBest}53.7 & \cellcolor{ColBest}87.3 & \cellcolor{ColBest}90.0 & \cellcolor{ColBest}88.6 & 100.0 \\
\logoGPT~GPT-5.6-Sol\badgeXHigh & \cellcolor{RankPurple!41}2 & \cellcolor{ColSecond}67.49 & 38.0 & 39.9 & \cellcolor{ColSecond}69.0 & \cellcolor{ColBest}53.1 & 43.6 & 50.7 & 47.3 & 16.0 & 51.7 & \cellcolor{ColSecond}84.0 & \cellcolor{ColSecond}86.7 & \cellcolor{ColSecond}85.4 & 98.6 \\
\logoGrok~Grok-4.6\badgeXHigh & \cellcolor{RankPurple!40}3 & 67.17 & 40.5 & 42.2 & 66.2 & \cellcolor{ColSecond}46.9 & 42.7 & 52.5 & 59.1 & 24.0 & 52.1 & 82.9 & 85.5 & 84.2 & 97.9 \\
\logoQwen~Qwen3.8-Max\badgeHigh & \cellcolor{RankPurple!39}4 & 66.24 & \cellcolor{ColSecond}42.5 & 37.6 & 65.8 & 45.4 & 38.6 & \cellcolor{ColSecond}56.1 & \cellcolor{ColBest}72.0 & 30.0 & \cellcolor{ColSecond}52.7 & 79.9 & 82.8 & 81.3 & 99.3 \\
\logoClaude~Claude-Opus-5\badgeHigh & \cellcolor{RankPurple!39}5 & 66.21 & 42.0 & \cellcolor{ColBest}46.2 & 66.4 & 43.8 & 44.0 & 52.0 & 65.6 & 16.0 & 52.6 & 79.8 & 82.9 & 81.4 & 100.0 \\
\logoGPT~GPT-5.6-Sol\badgeHigh & \cellcolor{RankPurple!38}6 & 65.73 & 34.5 & 37.6 & 65.0 & 40.8 & 37.8 & 52.2 & 63.4 & 26.0 & 49.8 & 82.5 & 85.3 & 83.9 & 99.7 \\
\logoGPT~GPT-5.6-Terra\badgeXHigh & \cellcolor{RankPurple!37}7 & 64.79 & 33.0 & 30.6 & 65.8 & 43.1 & 40.2 & 50.7 & 62.4 & 12.0 & 48.9 & 81.6 & 84.3 & 82.9 & 97.9 \\
\logoGPT~GPT-5.6-Terra\badgeHigh & \cellcolor{RankPurple!37}8 & 64.76 & 37.5 & 34.1 & 66.2 & 45.4 & 30.3 & 52.5 & 63.4 & 16.0 & 49.2 & 81.0 & 83.9 & 82.4 & 99.7 \\
\logoGPT~GPT-5.5\badgeHigh & \cellcolor{RankPurple!36}9 & 64.53 & 34.5 & 34.7 & 65.8 & 46.2 & 37.8 & 50.5 & 63.4 & 12.0 & 49.3 & 80.4 & 83.2 & 81.8 & 97.2 \\
\logoGPT~GPT-5.6-Sol & \cellcolor{RankPurple!35}10 & 64.09 & 31.0 & 37.0 & 67.1 & 37.7 & 40.2 & 50.7 & 61.3 & 22.0 & 49.5 & 79.1 & 82.0 & 80.6 & 99.7 \\
\logoGPT~GPT-5.6-Sol\badgeMed & \cellcolor{RankPurple!35}11 & 64.01 & 35.0 & 40.5 & 64.3 & 33.8 & 41.1 & 48.0 & 61.3 & 18.0 & 48.5 & 80.1 & 83.1 & 81.6 & 99.7 \\
\logoGPT~GPT-5.6-Luna\badgeHigh & \cellcolor{RankPurple!33}13 & 63.58 & 28.5 & 36.4 & 65.2 & 34.6 & 37.3 & 54.4 & 54.8 & 10.0 & 48.2 & 79.6 & 82.7 & 81.1 & 99.0 \\
\logoMeta~Muse-Spark-1.3\badgeXHigh & \cellcolor{RankPurple!33}14 & 63.40 & 35.0 & 36.4 & 64.1 & 35.4 & 40.2 & 51.2 & 62.4 & 20.0 & 48.9 & 78.3 & 81.2 & 79.7 & 99.0 \\
\logoGPT~GPT-5.6-Sol\badgeLow & \cellcolor{RankPurple!32}15 & 63.23 & 28.5 & 41.0 & 63.0 & 35.4 & 39.4 & 54.4 & 62.4 & 22.0 & 49.0 & 77.8 & 80.8 & 79.3 & 99.3 \\
\logoGrok~Grok-4.5\badgeHigh & \cellcolor{RankPurple!31}16 & 62.81 & 35.0 & 39.3 & 64.1 & 42.3 & 39.8 & 47.3 & 59.1 & 14.0 & 48.4 & 77.6 & 80.4 & 79.0 & 96.5 \\
\logoGPT~GPT-5.5\badgeLow & \cellcolor{RankPurple!31}17 & 62.80 & 35.0 & 33.5 & 61.8 & 42.3 & 44.8 & 54.4 & 55.9 & 26.0 & 49.6 & 76.0 & 79.0 & 77.5 & 99.0 \\
\logoGPT~GPT-5.5\badgeMed & \cellcolor{RankPurple!30}18 & 62.56 & 25.5 & 35.8 & 63.9 & 41.5 & 37.8 & 50.0 & 51.6 & 12.0 & 46.9 & 79.1 & 81.9 & 80.5 & 97.2 \\
\logoGPT~GPT-5.6-Terra\badgeMed & \cellcolor{RankPurple!29}19 & 62.43 & 36.0 & 34.1 & 62.4 & 39.2 & 43.2 & 47.3 & 55.9 & \cellcolor{ColSecond}34.0 & 48.2 & 77.0 & 80.1 & 78.5 & 100.0 \\
\logoClaude~Claude-Sonnet-4.6 & \cellcolor{RankPurple!29}20 & 62.37 & 30.5 & 37.0 & 63.9 & 43.8 & 43.2 & \cellcolor{ColBest}57.4 & 62.4 & 14.0 & 50.6 & 73.8 & 76.9 & 75.3 & 99.7 \\
\logoGemini~Gemini-3.1-Pro\badgeHigh & \cellcolor{RankPurple!28}21 & 62.23 & 30.0 & 44.5 & 65.2 & 43.1 & 41.1 & 55.1 & 63.4 & 20.0 & 51.1 & 72.9 & 76.1 & 74.5 & 97.9 \\
\logoGemini~Gemini-3.8-Flash\badgeHigh & \cellcolor{RankPurple!27}22 & 62.08 & 41.5 & 35.8 & 65.4 & 37.7 & 42.7 & 44.9 & 51.6 & 18.0 & 48.4 & 76.0 & 78.8 & 77.4 & 95.1 \\
\logoGPT~GPT-5.6-Luna\badgeMed & \cellcolor{RankPurple!27}23 & 61.62 & 23.0 & 31.8 & 62.2 & 35.4 & 41.9 & 52.5 & 53.8 & 14.0 & 46.5 & 77.2 & 80.5 & 78.8 & 100.0 \\
\logoClaude~Claude-Sonnet-5 & \cellcolor{RankPurple!26}24 & 61.24 & 30.0 & 39.9 & 62.4 & 36.2 & 40.7 & 54.9 & 64.5 & 18.0 & 49.2 & 72.9 & 76.3 & 74.6 & 99.7 \\
\logoGPT~GPT-5.2\badgeHigh & \cellcolor{RankPurple!25}25 & 61.08 & 39.0 & 38.2 & 60.9 & 46.2 & 32.4 & 51.2 & 59.1 & 12.0 & 47.9 & 74.4 & 77.2 & 75.8 & 99.0 \\
\logoGPT~GPT-5.6-Terra\badgeLow & \cellcolor{RankPurple!25}26 & 61.07 & 27.0 & 36.4 & 62.4 & 38.5 & 40.2 & 49.0 & 57.0 & 24.0 & 47.1 & 75.2 & 78.4 & 76.8 & 99.7 \\
\logoClaude~Claude-Sonnet-5\badgeHigh & \cellcolor{RankPurple!24}27 & 60.63 & 29.0 & 37.0 & 65.4 & 28.5 & 39.8 & 54.7 & 54.8 & 10.0 & 48.3 & 72.7 & 76.1 & 74.4 & 100.0 \\
\logoClaude~Claude-Sonnet-4.6\badgeHigh & \cellcolor{RankPurple!23}28 & 60.53 & 35.5 & 32.4 & 62.8 & 34.6 & 36.9 & 52.9 & 58.1 & 12.0 & 47.7 & 73.4 & 76.5 & 74.9 & 97.9 \\
\logoGPT~GPT-5.6-Luna\badgeLow & \cellcolor{RankPurple!22}29 & 60.33 & 20.5 & 35.8 & 60.5 & 34.6 & 43.2 & 51.2 & 67.7 & 18.0 & 46.8 & 73.9 & 77.3 & 75.6 & 100.0 \\
\logoClaude~Claude-Opus-4.7 & \cellcolor{RankPurple!22}30 & 60.32 & 36.0 & 41.0 & 61.5 & 34.6 & 42.3 & 54.9 & 57.0 & 8.0 & 49.2 & 70.9 & 74.3 & 72.6 & 97.6 \\
\logoClaude~Claude-Opus-4.6 & \cellcolor{RankPurple!21}31 & 60.23 & 28.0 & 39.3 & 63.7 & 30.0 & 46.1 & 48.0 & 50.5 & 24.0 & 47.5 & 73.0 & 75.9 & 74.5 & 99.0 \\
\logoMeta~Muse-Spark-1.2\badgeXHigh & \cellcolor{RankPurple!20}32 & 59.90 & 33.5 & 32.4 & 62.8 & 36.9 & 34.0 & 47.5 & 62.4 & 10.0 & 46.2 & 74.1 & 76.7 & 75.4 & 93.8 \\
\logoClaude~Claude-Opus-4.7\badgeHigh & \cellcolor{RankPurple!20}33 & 59.17 & 33.5 & 42.2 & 61.5 & 35.4 & 38.2 & 53.7 & 55.9 & 14.0 & 48.3 & 69.3 & 72.9 & 71.1 & 96.9 \\
\logoClaude~Claude-Opus-4.8 & \cellcolor{RankPurple!19}34 & 59.07 & 30.5 & 37.0 & 62.0 & 35.4 & 41.9 & 48.8 & 57.0 & 18.0 & 47.2 & 70.4 & 74.1 & 72.2 & 100.0 \\
\logoGemini~Gemini-3-Flash\badgeHigh & \cellcolor{RankPurple!18}35 & 58.91 & 31.5 & 38.7 & 66.4 & 40.8 & 39.0 & 52.5 & 58.1 & 18.0 & 49.6 & 67.3 & 70.7 & 69.0 & 100.0 \\
\logoClaude~Claude-Opus-4.8\badgeHigh & \cellcolor{RankPurple!18}36 & 58.46 & 29.0 & 33.5 & 60.9 & 30.0 & 41.1 & 51.7 & 54.8 & 16.0 & 46.4 & 69.9 & 73.9 & 71.9 & 100.0 \\
\logoGPT~GPT-5.2 & \cellcolor{RankPurple!17}37 & 57.95 & 30.0 & 41.0 & 60.9 & 31.5 & 40.2 & 50.7 & 50.5 & 12.0 & 46.7 & 68.7 & 72.2 & 70.4 & 99.7 \\
\logoGPT~GPT-5.5 & \cellcolor{RankPurple!16}38 & 57.03 & 21.5 & 24.3 & 59.8 & 30.8 & 34.9 & 46.3 & 59.1 & 14.0 & 42.6 & 71.8 & 75.4 & 73.6 & 99.3 \\
\logoGPT~GPT-5.4 & \cellcolor{RankPurple!16}39 & 56.80 & 23.5 & 41.6 & 64.8 & 43.1 & 32.8 & 48.5 & 50.5 & 16.0 & 46.6 & 66.3 & 69.6 & 68.0 & 97.6 \\
\logoGPT~GPT-5.4-Mini & \cellcolor{RankPurple!15}40 & 56.51 & 19.5 & \cellcolor{ColSecond}45.1 & 60.9 & 23.8 & 40.7 & 52.2 & 63.4 & 14.0 & 46.5 & 65.7 & 69.4 & 67.5 & 100.0 \\
\logoGLM~GLM-5V-Turbo & \cellcolor{RankPurple!14}41 & 56.47 & 25.5 & 42.2 & 61.5 & 36.9 & 32.8 & 52.9 & 59.1 & 12.0 & 46.8 & 65.4 & 68.7 & 67.0 & 94.1 \\
\logoGrok~Grok-4.3\badgeHigh & \cellcolor{RankPurple!12}44 & 54.08 & 22.0 & 36.4 & 57.5 & 32.3 & 41.1 & 51.0 & 52.7 & 12.0 & 44.7 & 62.8 & 65.9 & 64.3 & 94.4 \\
\logoSeed~Seed-2.0-Mini & \cellcolor{RankPurple!10}47 & 52.24 & 17.0 & 31.8 & 61.5 & 20.8 & 31.1 & 52.9 & 61.3 & 4.0 & 43.4 & 59.8 & 64.0 & 61.9 & 97.2 \\
\logoQwen~Qwen3.6-Plus & \cellcolor{RankPurple!10}48 & 51.28 & 20.5 & 31.8 & 54.1 & 30.0 & 29.9 & 48.5 & 61.3 & 12.0 & 41.4 & 60.8 & 63.7 & 62.2 & 88.5 \\
\logoGrok~Grok-4.3 & \cellcolor{RankPurple!9}49 & 50.59 & 13.0 & 29.5 & 59.0 & 30.0 & 40.2 & 52.5 & \cellcolor{ColSecond}68.8 & 12.0 & 44.4 & 55.4 & 59.0 & 57.2 & 99.0 \\
\logoSeed~Seed-2.0-Lite & \cellcolor{RankPurple!8}50 & 49.47 & 10.5 & 33.5 & 57.9 & 30.0 & 32.0 & 48.5 & 52.7 & 8.0 & 41.3 & 56.2 & 60.6 & 58.4 & 92.7%
\\
\multicolumn{4}{l}{\cellcolor{ColOpenSec}\textit{Open-weight Models}} &
\multicolumn{1}{@{}c@{}}{\cellcolor{ColOpenSec!90}} &
\multicolumn{1}{@{}c@{}}{\cellcolor{ColOpenSec!81}} &
\multicolumn{1}{@{}c@{}}{\cellcolor{ColOpenSec!72}} &
\multicolumn{1}{@{}c@{}}{\cellcolor{ColOpenSec!63}} &
\multicolumn{1}{@{}c@{}}{\cellcolor{ColOpenSec!54}} &
\multicolumn{1}{@{}c@{}}{\cellcolor{ColOpenSec!45}} &
\multicolumn{1}{@{}c@{}}{\cellcolor{ColOpenSec!36}} &
\multicolumn{1}{@{}c@{}}{\cellcolor{ColOpenSec!27}} &
\multicolumn{1}{@{}c@{}}{\cellcolor{ColOpenSec!18}} &
\multicolumn{1}{@{}c@{}}{\cellcolor{ColOpenSec!10}} &
\multicolumn{1}{@{}c@{}}{\cellcolor{ColOpenSec!4}} &
\multicolumn{1}{@{}c@{}}{} \\
\logoGLM~GLM-5.3-Flash\badgeXHigh & \cellcolor{RankPurple!34}12 & 63.96 & 35.0 & 38.2 & 64.5 & 46.2 & \cellcolor{ColSecond}46.9 & 51.7 & 54.8 & 28.0 & 50.8 & 77.0 & 80.3 & 78.7 & 99.3 \\
\logoKimi~Kimi-K2.5 & \cellcolor{RankPurple!14}42 & 55.91 & 24.0 & 32.9 & 59.2 & 33.8 & 36.9 & 47.5 & 58.1 & 6.0 & 44.0 & 67.7 & 70.8 & 69.2 & 95.8 \\
\logoMiniMax~MiniMax-M3 & \cellcolor{RankPurple!13}43 & 55.29 & 20.0 & 35.8 & 61.5 & 30.8 & 40.2 & 49.5 & 62.4 & 14.0 & 45.6 & 64.3 & 67.6 & 65.9 & 97.6 \\
\logoQwen~Qwen3.5-122B & \cellcolor{RankPurple!12}45 & 53.50 & 16.5 & 32.9 & 63.9 & 25.4 & 29.0 & 52.9 & 57.0 & 8.0 & 44.1 & 62.0 & 65.6 & 63.8 & 97.6 \\
\logoQwen~Qwen3.5-397B & \cellcolor{RankPurple!11}46 & 52.75 & 16.0 & 32.9 & 60.9 & 29.2 & 38.2 & 45.1 & 57.0 & 10.0 & 43.0 & 61.9 & 65.2 & 63.5 & 97.2 \\
\logoQwen~Qwen3.5-27B & \cellcolor{RankPurple!8}51 & 47.50 & 13.5 & 27.2 & 54.5 & 26.2 & 30.7 & 45.3 & 49.5 & 4.0 & 38.6 & 55.8 & 58.9 & 57.3 & 84.7%
\\
\bottomrule
\end{tabular}%
}
\end{table}

\section{Leaderboard Uncertainty}
\label{app:uncertainty}

We quantify the gap between GPT-6-Astra-\texttt{high} and
GPT-5.6-Sol-\texttt{xhigh} with a paired scene bootstrap.
Each of 10,000 repetitions samples the 288 scene IDs with replacement, then
recomputes DV from retained and original-correct question counts, LS from the
scene mean, and \SqrtMean{} from the two resampled axes.

\begin{table}[h]
\centering
\caption{Paired scene-bootstrap intervals for the top two configurations.
Brackets give 95\% percentile intervals, and the final column is the fraction
of bootstrap repetitions with a positive Astra-minus-Sol difference.}
\label{tab:two_axis_uncertainty}
\setlength{\tabcolsep}{4pt}
\resizebox{\linewidth}{!}{%
\begin{tabular}{lcccc}
\toprule
Metric & Astra high & Sol xhigh & Difference & $\Pr_{\mathrm{boot}}(\Delta>0)$ \\
\midrule
Dual VQA & 53.69 [51.24, 56.18] & 51.72 [49.31, 54.12] & +1.97 [-0.91, +4.79] & 0.903 \\
Latent Similarity & 88.62 [88.32, 88.92] & 85.36 [84.00, 86.46] & +3.27 [+2.20, +4.60] & 1.000 \\
Overall & 70.07 [68.65, 71.51] & 67.49 [65.78, 69.10] & +2.58 [+0.72, +4.45] & 0.997%
\\
\bottomrule
\end{tabular}
}
\end{table}

The \SqrtMean{} lead is 2.58 points with a 95\% interval of
$[0.72,4.45]$.
The LS advantage is likewise stable, while the 1.97-point DV difference has an
interval that crosses zero.
We therefore treat Astra as the \SqrtMean{} and LS leader but do not claim a
statistically significant DV advantage over Sol-\texttt{xhigh}.

\section{Derivation of the Square-Root Mean}
\label{sec:appendix_sqrt_mean}

For configuration $i$, let the set of evaluation axes be
$\mathcal{A}=\{\mathrm{DV},\mathrm{LS}\}$.
Writing $r_i^a=\sqrt{s_i^a}$ and
$\bar r_i=|\mathcal{A}|^{-1}\sum_{a\in\mathcal{A}}r_i^a$,
Equation~\eqref{eq:sqrt_mean} defines $\bar s_i=\bar r_i^2$.
Using the population variance across axes,
\begin{align}
\mathrm{Var}_{a\in\mathcal{A}}(r_i^a)
&= \frac{1}{|\mathcal{A}|}\sum_{a\in\mathcal{A}}s_i^a-\bar s_i,\\
\bar s_i
&= \frac{\mathrm{DV}_i+\mathrm{LS}_i
+2\sqrt{\mathrm{DV}_i\mathrm{LS}_i}}{4}.
\end{align}
Hence the square-root mean is the arithmetic mean corrected by
$\mathrm{Var}(\sqrt{s_i^a})$.
It lies between the geometric and arithmetic means, increases whenever either
axis improves, and remains nonzero when only one axis is zero.
As a protective measure, the implementation clips each axis score to
$\max(0,\,s_i^a)$ before taking the square root; no evaluated
configuration produces a negative score.
The choice does not determine the headline ordering. GPT-6 Astra, Sol-xhigh,
and Grok-4.6-xhigh remain the top three under arithmetic and geometric aggregation,
and the full 51-model ranks correlate with the square-root ranks at
$\rho{=}0.999$ and $0.998$, respectively.

\section{Dual VQA Protocol}
\label{sec:appendix_dual_vqa}

Dual VQA uses the judge \texttt{gpt-5.4-mini}.
For every VSI-Bench question $q$, we sample 16 frames from the original video
and 16 from the reconstructed camera render, ask the same shortest-answer
prompt, and score both answers against the VSI-Bench target.
Original-video answers are queried once and shared across every model run.
The prompt instructs the judge to use only the provided frames and return a
number, short phrase, or option letter without explanation.
We set \texttt{max\_completion\_tokens=32} and \texttt{temperature=0}, and
retry transient API failures up to six times.
All reported runs use the same \texttt{gpt-5.4-mini} API alias.

Let $o_q$ and $r_{i,q}$ indicate whether the original and reconstruction
answers are correct for configuration $i$.
We report conditional retention
\begin{equation}
\mathrm{DV}_i =
\frac{\sum_q o_q r_{i,q}}{\sum_q o_q}.
\end{equation}
The denominator contains the 1,827 questions the judge answers correctly from
the original clips, out of 5,130 total questions.
The judge-correct denominators for object count, absolute distance, object size,
room size, relative distance, relative direction, route planning, and appearance
order are respectively 200, 173, 532, 130, 241, 408, 93, and 50.
This control prevents judge failures on the source video from being attributed
to a reconstruction.
Because reconstructions consist of untextured Blender primitives whose
appearance differs from real video, the judge may find reconstruction frames
harder to parse. A lower judge accuracy on reconstruction frames
therefore reflects information lost during reconstruction, not a flaw
in the evaluation.
Results are also reported by VSI-Bench task, with relative-direction easy,
medium, and hard subsets micro-aggregated into one column.

\section{Latent Similarity with V-JEPA~2.1}
\label{sec:appendix_vjepa}

BVB compares each original video and camera render with a frozen
V-JEPA~2.1 ViT-G encoder
(\texttt{apiantonio/vjepa2.1-vit-gigantic-384})~\citep{bardes2024vjepa,assran2025vjepa2,murlabadia2026vjepa21}.
The encoder is never fine-tuned on BVB.
This makes the metric independent of the evaluated agent and avoids training a
scoring head on benchmark reconstructions.

\paragraph{Paired clips.}
We render $T{=}64$ RGB frames uniformly across the reconstruction's Blender camera
timeline with EEVEE at 512-pixel resolution and sample 64 frames uniformly from
the original mp4.
Both clips cover their full temporal extent without assuming framewise
registration.

\paragraph{Encoding and scores.}
Let $z_{\mathrm{orig}},z_{\mathrm{rend}}\in\mathbb{R}^{N\times D}$ denote the
last-layer token sequences.
We average all tokens for a global clip representation and reshape tokens
onto a $T_g\times H\times W$ grid for a temporally pooled layout map.
With 64 input frames, patch size 16, and tubelet size 2, the ViT-G encoder
produces a $32\times 32\times 32$ token grid ($N{=}32{,}768$ tokens of
dimension $D{=}1664$):
\begin{align}
\bar{z}^{\mathrm{mot}}_{c}
&=\mathrm{mean}_{i=1}^{N}(z_{c,i}),\\
\bar{z}^{\mathrm{lay}}_{c}
&=\mathrm{mean}_{t=1}^{T_g}
\big(\mathrm{reshape}(z_c;T_g,H,W,D)_{t,:,:,:}\big),\\
\mathrm{motion\_sim}
&=\cos\big(\bar{z}^{\mathrm{mot}}_{\mathrm{orig}},
\bar{z}^{\mathrm{mot}}_{\mathrm{rend}}\big),\\
\mathrm{layout\_sim}
&=\frac{1}{HW}\sum_{h,w}
\cos\big(\bar{z}^{\mathrm{lay}}_{\mathrm{orig}}[h,w],
\bar{z}^{\mathrm{lay}}_{\mathrm{rend}}[h,w]\big),\\
\mathrm{LS}
&=\tfrac{1}{2}
\big(\mathrm{layout\_sim}+\mathrm{motion\_sim}\big).
\end{align}
Features are discarded after scoring.
The release retains per-scene JSONL records and one run-level summary.

\section{Additional Score Diagnostics}
\label{app:score_diagnostics}

\begin{figure}[H]
  \centering
  \begin{minipage}[b]{0.57\linewidth}
    \centering
    \includegraphics[width=\linewidth]{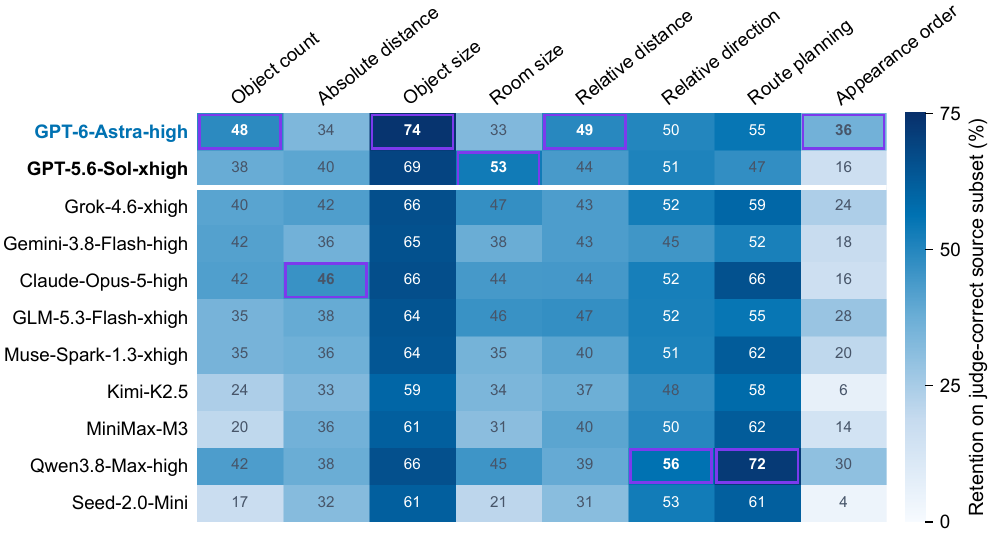}\\[-1pt]
    {\footnotesize (a) Dual VQA retention by task}
  \end{minipage}\hfill
  \begin{minipage}[b]{0.41\linewidth}
    \centering
    \includegraphics[width=\linewidth]{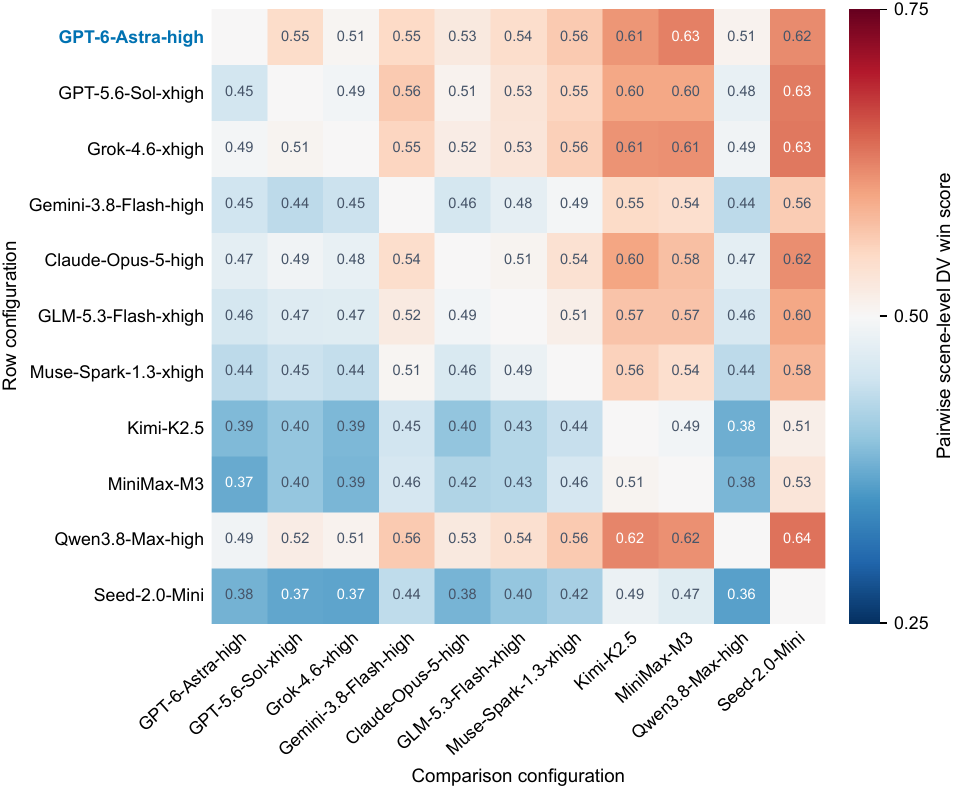}\\[-1pt]
    {\footnotesize (b) Pairwise scene-level DV wins}
  \end{minipage}
  \caption{\textbf{Dual VQA difficulty is structured, and aggregate ranks do not
  reflect scene-level ties.}
  (a) Object size and route planning are retained more reliably than appearance
  order and object count. Purple boxes mark the best shown value in each task.
  (b) Each off-diagonal cell is the row configuration's scene-level DV win
  score against the column configuration over the same 278 scenes.
  Ties contribute one half.}
  \label{fig:appendix_dv_task_heatmap}
  \label{fig:appendix_pairwise_dv}
\end{figure}

\begin{figure}[H]
  \centering
  \includegraphics[width=\linewidth]{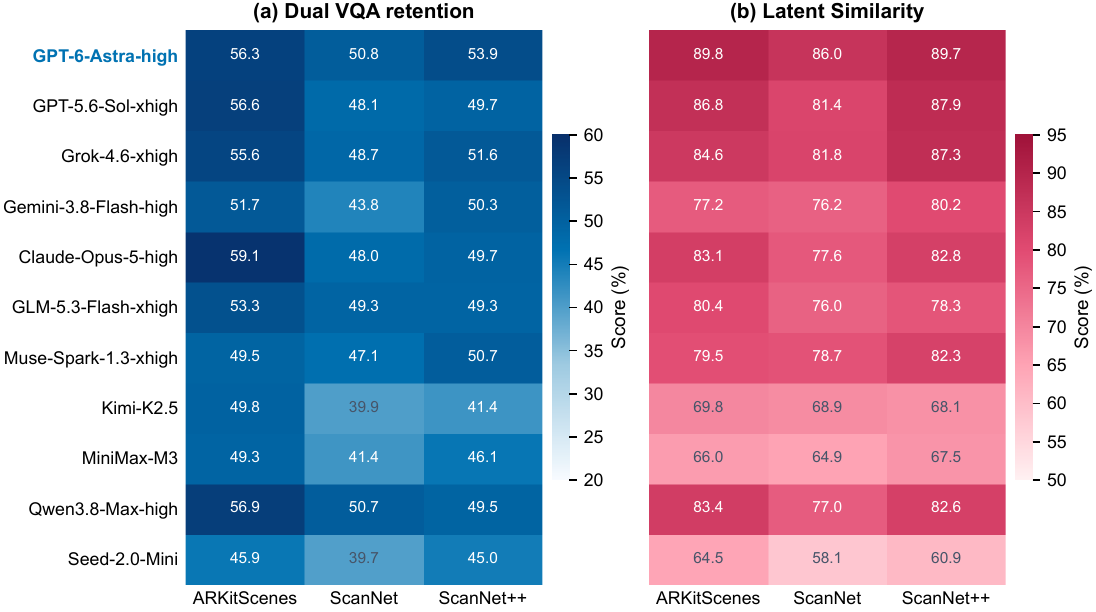}
  \caption{\textbf{Source collection shifts both evaluation axes.}
  Every shown configuration scores higher on ARKitScenes than ScanNet for both
  DV and LS.
  DV is micro-averaged over source-correct questions within each collection,
  while LS is averaged over its scene records.}
  \label{fig:appendix_source_profiles}
\end{figure}

\begin{figure}[H]
  \centering
  \includegraphics[width=\linewidth]{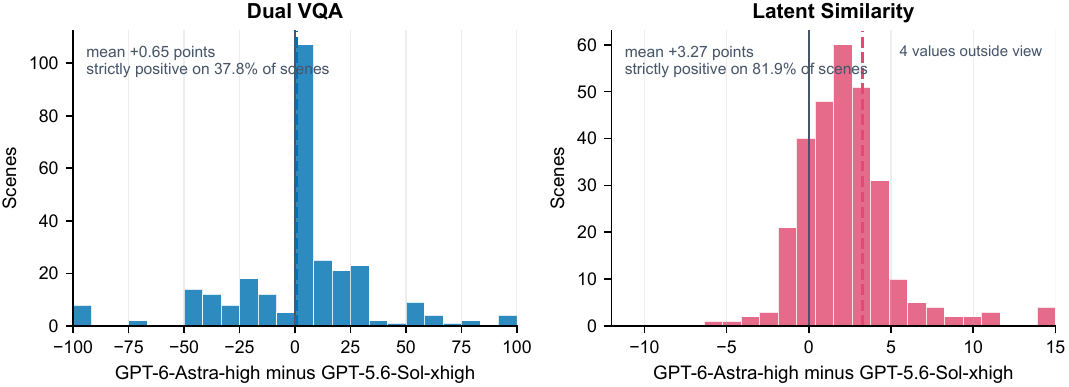}
  \caption{\textbf{Top-line gains are not uniform across scenes.}
  DV contains many identical scores and both positive and negative scene differences.
  LS is positive on most scenes.
  Four LS values beyond the displayed range correspond to Sol render failures
  scored as zero under the leaderboard policy.}
  \label{fig:appendix_scene_differences}
\end{figure}

These diagnostics use the same fixed source-correct question set, missing
output handling, and per-scene LS records as the leaderboard.
They show variation that a single run-level aggregate does not capture.
Figure~\ref{fig:appendix_dv_task_heatmap} enlarges the task profile from the
main text and reports scene-level pairwise wins.
Figures~\ref{fig:appendix_source_profiles}--\ref{fig:appendix_scene_differences}
then compare source collections and paired score differences.

Panel (b) of Figure~\ref{fig:appendix_pairwise_dv} gives the top pair a
scene-level win score of 0.55, considerably less decisive than a rank alone
suggests.
This agrees with the paired-bootstrap result that the top-two DV difference is
not statistically significant.
Table~\ref{tab:appendix_dv_task_distribution} gives the exact distribution
summary behind the task-level plots.

\begin{table}[H]
\centering
\caption{\textbf{Task difficulty and discrimination differ across BVB tasks.}
Statistics summarize task-level DV over all 51 configurations.
Object size has the highest mean, while object count has the widest range.}
\label{tab:appendix_dv_task_distribution}
\setlength{\tabcolsep}{9pt}
\begin{tabular}{lccc}
\toprule
Task & Mean & Interquartile range & Full range \\
\midrule
Object count & 29.3 & $[23.2, 35.0]$ & $[10.5, 48.5]$ \\
Absolute distance & 36.5 & $[33.5, 39.6]$ & $[24.3, 46.2]$ \\
Object size & 63.0 & $[61.2, 65.2]$ & $[54.1, 73.7]$ \\
Room size & 36.5 & $[31.2, 42.3]$ & $[20.8, 53.1]$ \\
Relative distance & 39.0 & $[37.1, 41.9]$ & $[29.0, 49.0]$ \\
Relative direction & 51.0 & $[48.7, 52.9]$ & $[44.9, 57.4]$ \\
Route planning & 58.5 & $[54.8, 62.4]$ & $[47.3, 72.0]$ \\
Appearance order & 16.1 & $[12.0, 19.0]$ & $[4.0, 36.0]$%
\\
\bottomrule
\end{tabular}
\end{table}

\paragraph{Recovery on judge-failed source questions.}
Retention conditions on the questions the judge answers correctly from the
source, and we examine recovery on questions outside this subset.
Restricted to scenes with a successful render, the judge recovers on average
22.6\% of the judge-failed multiple-choice questions from the render, and no
configuration exceeds 25.1\%.
Both values sit below the 27.9\% chance level obtained by blending the two-,
three-, and four-option formats over this question pool.
Recovery on these questions remains below the nominal chance baseline
across all evaluated configurations, indicating that the judge's errors
on these questions are systematic rather than random and that the
reconstruction does not provide additional signal to correct them.
On the numeric tasks the recovery rate instead ranges from 14.5\% to 19.9\%
and correlates with retention at Spearman 0.60 across the 51 configurations.

The source profile in Figure~\ref{fig:appendix_source_profiles} shows a
domain effect shared across otherwise different configurations.
Because the collections differ in capture style, scene composition, and
question mix, these differences show sensitivity to the source domain rather
than any single dataset factor.

The unweighted mean scene-level DV difference in
Figure~\ref{fig:appendix_scene_differences} is $+0.65$ points, whereas the
pooled question-level leaderboard difference is $+1.97$ points.
Scene-level DV denominators vary with the number of source-correct questions,
so the pooled score and paired scene summary answer different questions.

\section{Effort and Runtime Details}
\label{app:effort_details}

Table~\ref{tab:appendix_effort_monotonicity} reports how consistently each
metric changes as provider-defined reasoning effort increases.

\begin{table}[H]
\centering
\caption{\textbf{Latent Similarity responds most monotonically to effort.}
Entries are within-family Spearman correlations between ordered effort settings
and each score.}
\label{tab:appendix_effort_monotonicity}
\setlength{\tabcolsep}{9pt}
\begin{tabular}{lcccc}
\toprule
Model family & Settings & \abDV{} & \abLS{} & \SqrtMean{} \\
\midrule
GPT-5.5 & 4 & $+0.40$ & $+1.00$ & $+0.80$ \\
GPT-5.6 Luna & 3 & $+0.50$ & $+1.00$ & $+1.00$ \\
GPT-5.6 Sol & 5 & $+0.60$ & $+0.90$ & $+0.70$ \\
GPT-5.6 Terra & 4 & $+0.80$ & $+1.00$ & $+1.00$ \\
\midrule
Mean & & $\mathbf{+0.57}$ & $\mathbf{+0.97}$ & $\mathbf{+0.88}$%
\\
\bottomrule
\end{tabular}
\end{table}

\begin{figure}[H]
  \centering
  \begin{minipage}[b]{0.58\linewidth}
    \centering
    \includegraphics[width=\linewidth]{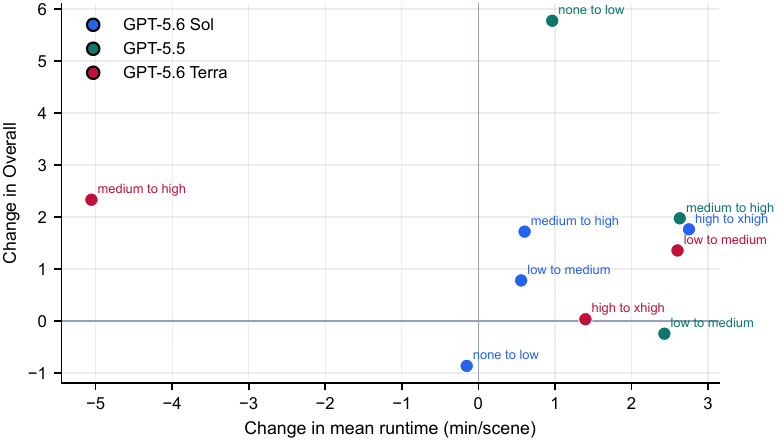}
    \caption{\textbf{Additional runtime is neither necessary nor sufficient for
    a \SqrtMean{} gain.}
    Each point compares adjacent effort settings within one provider family.}
    \label{fig:appendix_effort_marginal}
  \end{minipage}\hfill
  \begin{minipage}[b]{0.40\linewidth}
    \centering
    \includegraphics[width=\linewidth]{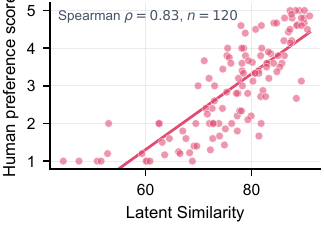}
    \caption{\textbf{Latent Similarity tracks blind visual preference.}
    Each point is one scene-model pair from the five-model study
    (Spearman $\rho{=}0.83$).}
    \label{fig:human_metric_calibration}
  \end{minipage}
\end{figure}

Figure~\ref{fig:appendix_effort_marginal} separates the runtime and score change
for each adjacent effort step.
Figure~\ref{fig:human_metric_calibration} reports the human calibration used in
the ranking study below.

The pattern supports treating effort as a provider-specific control rather than
a common compute scale.
It also explains why the main-text runtime scatter does not form a single
quality curve.

\section{Human Blind Ranking Study}
\label{app:human_study}

Fifteen raters each saw nine of a fixed pool of 24 scenes balanced across
ARKitScenes, ScanNet, and ScanNet++.
For every scene, raters viewed the source clip beside five anonymized
reconstructions and ranked them on object identity, spatial layout, camera
path, and appearance order.
The resulting 135 judgments per configuration yield mean ranks of 1.47, 2.23,
3.03, 3.70, and 4.58.
Table~\ref{tab:human_rank} reports these model-level results in the main text.

At the scene-model level, Figure~\ref{fig:human_metric_calibration} shows that
human preference correlates with Latent Similarity at Spearman $\rho{=}0.83$
over 120 observations.
Dual VQA has lower correlation ($\rho{=}0.11$, $n{=}110$), consistent with its
different role as a semantic retention measure.
Its sample is smaller because 10 scene-model cases contain no judge-correct
source question and therefore have no defined conditional retention.
These observations are clustered within 24 scenes and five configurations, so the
correlations are descriptive rather than based on 120 independent samples.
The study supports the perceptual metric and the ordering of the five sampled
configurations.
The gap reflects a difference in task format.
Ranking five reconstructions side by side is an inherently visual comparison,
so raters weight overall appearance even when the criteria name factual
elements.
Dual VQA scores discrete questions independently and captures a different
axis of reconstruction quality.

\begin{figure}[H]
  \centering
  \includegraphics[width=0.8\linewidth]{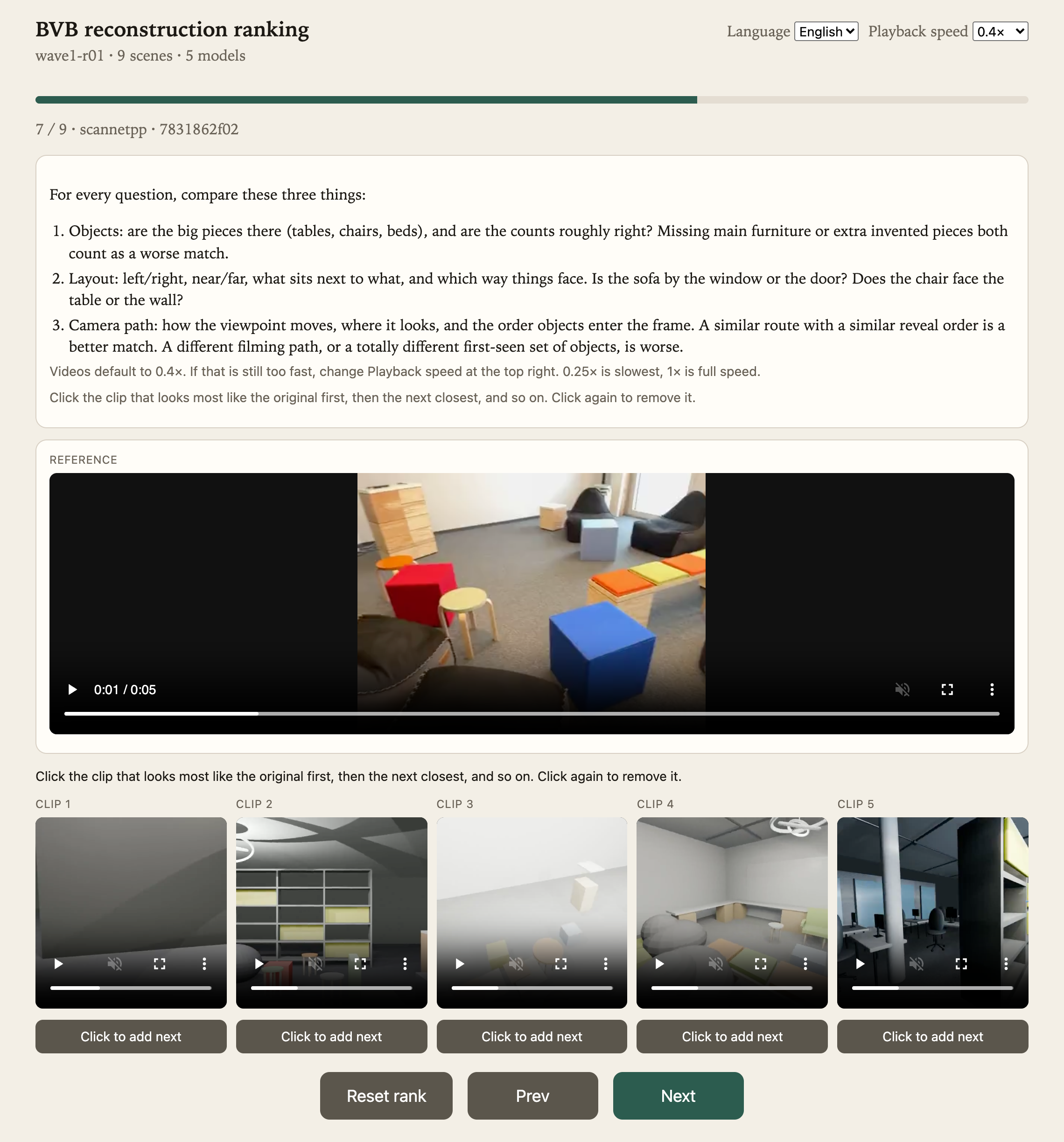}
  \caption{\textbf{Blind-ranking interface.}
  The reference video appears above five anonymized candidate reconstructions.
  Model identities were hidden, candidate order was randomized, and the form
  provided English and Chinese instructions.}
  \label{fig:human_study_interface}
\end{figure}

\section{Cost Frontier}
\label{app:cost_frontier}

For each configuration, agent spend is the mean Stage-1 API cost per scene from
provider usage accounting.
It excludes the Dual VQA judge and frozen V-JEPA encoder, so it measures the
cost of producing the reconstruction, not of scoring it.
Across 51 configurations, spend ranges from \$0.024 to \$2.157 per scene.

\begin{figure}[H]
  \centering
  \includegraphics[width=\linewidth]{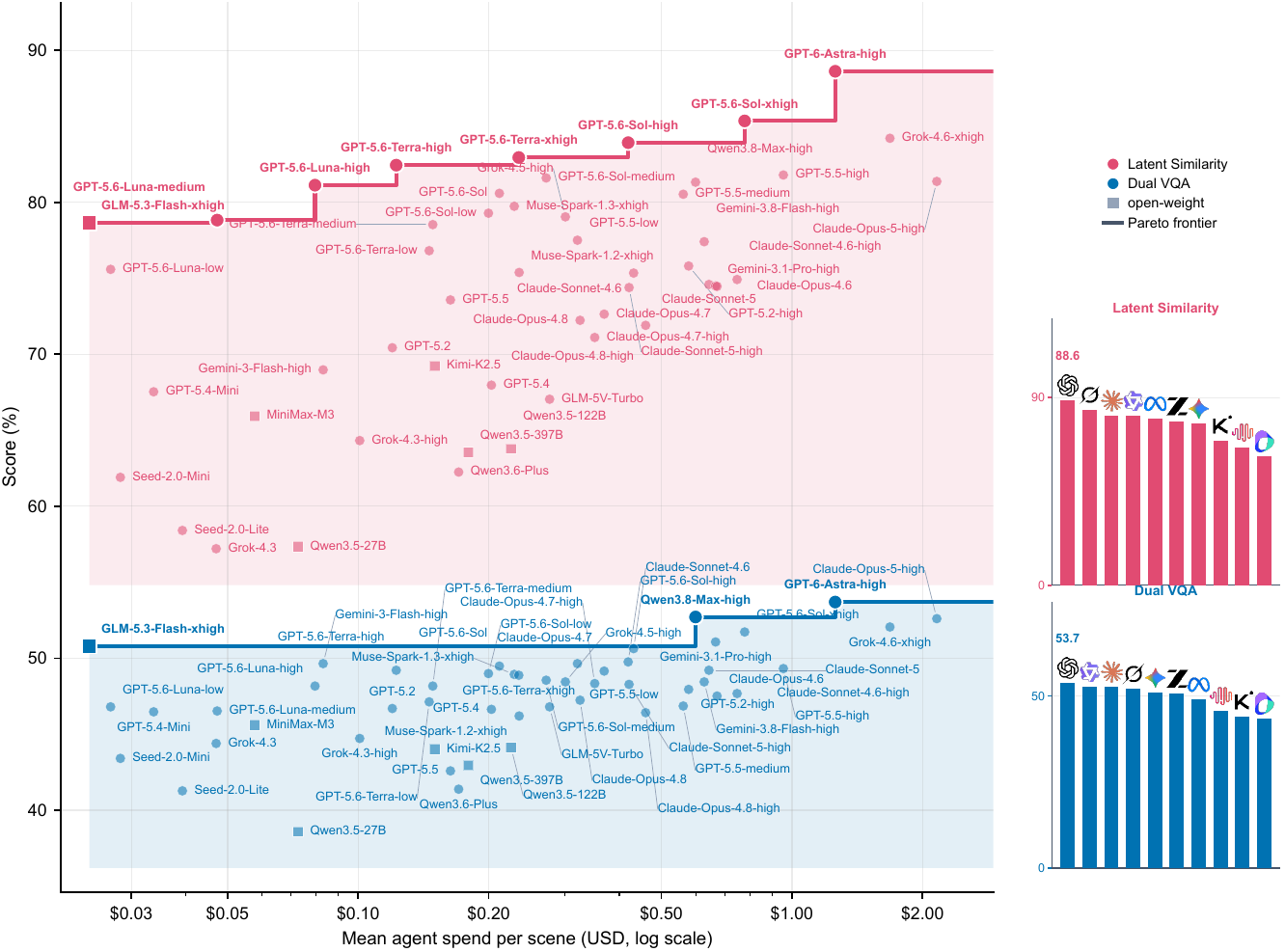}
  \caption{\textbf{Semantic and perceptual quality have different cost
  frontiers.}
  Each staircase contains configurations that no cheaper run outperforms on
  that axis, and the right panels show the best configuration from each model
  family.}
  \label{fig:cost_frontier_all}
\end{figure}

Three configurations lie on the DV frontier and eight on the LS frontier.
Both start with GLM-5.3-Flash and end with GPT-6 Astra; the DV frontier
passes through Qwen3.8-Max, while the LS frontier includes Luna, Terra, and Sol.
Astra reaches 53.7 \abDV{} and 88.6 \abLS{} at \$1.258 per scene.
Sol-\texttt{xhigh} reaches 51.7 and 85.4 at \$0.778, while
GLM-5.3-Flash-\texttt{xhigh} reaches 50.8 and 78.7 at \$0.024.
GLM-5.3-Flash is the only open-weight configuration on either measured frontier.

\section{Limitations}
\label{app:limitations}

BVB covers indoor egocentric videos from three capture sources, and broader
environments, outdoor scenes, and interactive editing remain future work.
Performance also depends on coding ability and familiarity with Blender, so
BVB evaluates video understanding through an agent's ability to express it
programmatically.
Both axes evaluate the rendered video, not the underlying 3D geometry, which matches
the source-conditioned setting where aligned ground-truth geometry is
unavailable.
As in other judge-based evaluations, Dual VQA depends on a VLM judge.
Conditioning on the judge-correct source subset and keeping the judge fixed
across all configurations mitigate this dependence.
The blind ranking study covers five configurations, and extending it to newer
models is straightforward under the released protocol.
Finally, the shared per-scene cost ceiling keeps comparisons uniform and
affordable, though individual configurations might improve further under larger
budgets.

\section{Qualitative Reconstruction Cases}
\label{app:qualitative}

The full candidate pool contains four scenes from each source.
Figure~\ref{fig:qualitative_cross_model} shows Candidate 12 in the main text,
and the remaining eleven appear below.
Every row uses the same six normalized clip times.
From top to bottom, each candidate shows the reference, GPT-6 Astra,
GPT-5.6 Sol, Grok, Gemini, Claude, and Qwen.
The five non-Astra configurations are exactly those used in the human blind
study.
These candidates support visual selection and do not estimate failure
prevalence.

\begin{figure}[H]
  \centering
  \includegraphics[width=0.80\linewidth]{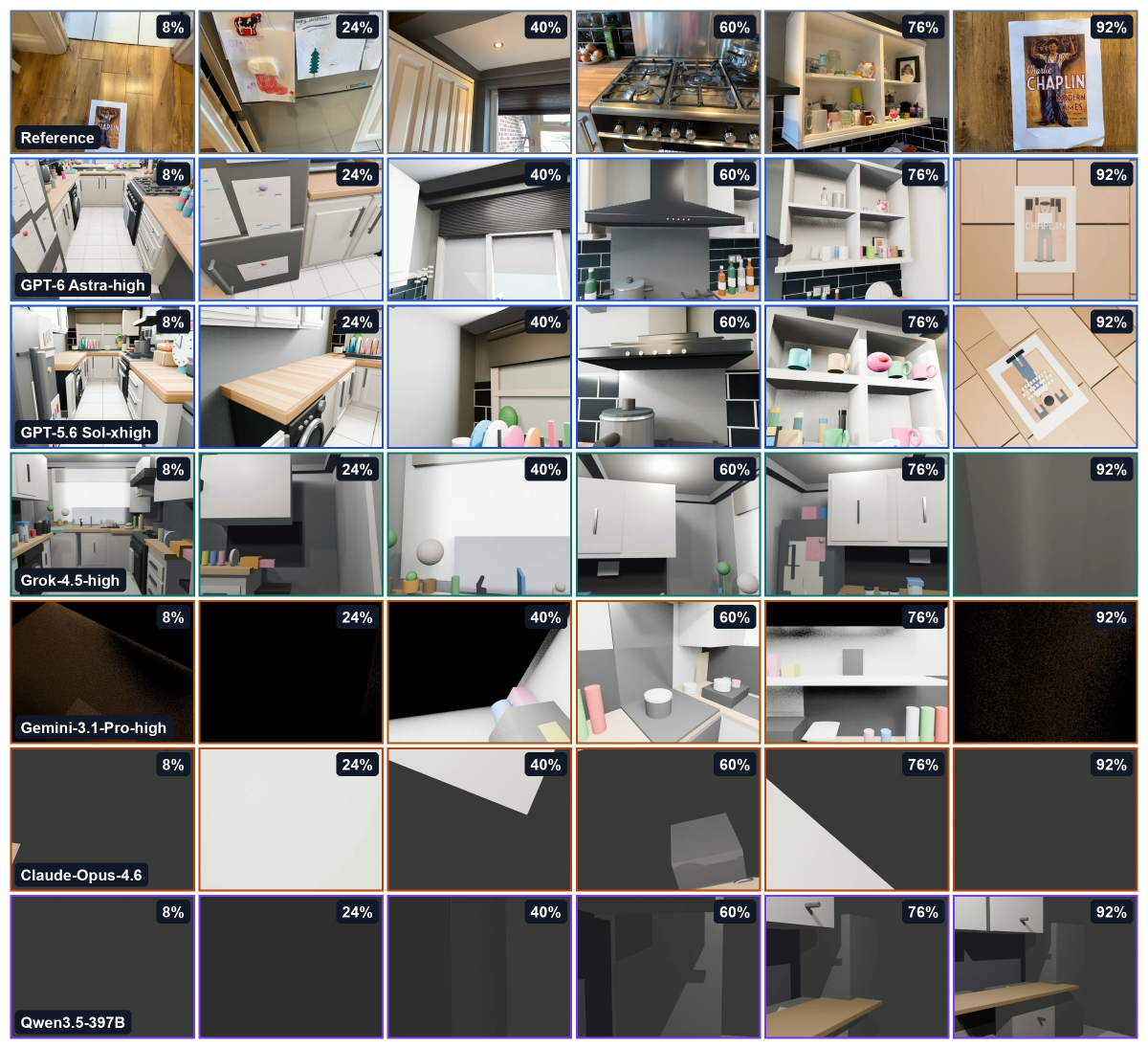}
  \caption{\textbf{Candidate 1: cluttered kitchen.}
  ARKitScenes scene 42446049 at six matched timestamps.}
  \label{fig:qualitative_candidate_01}
\end{figure}

\begin{figure}[H]
  \centering
  \includegraphics[width=0.80\linewidth]{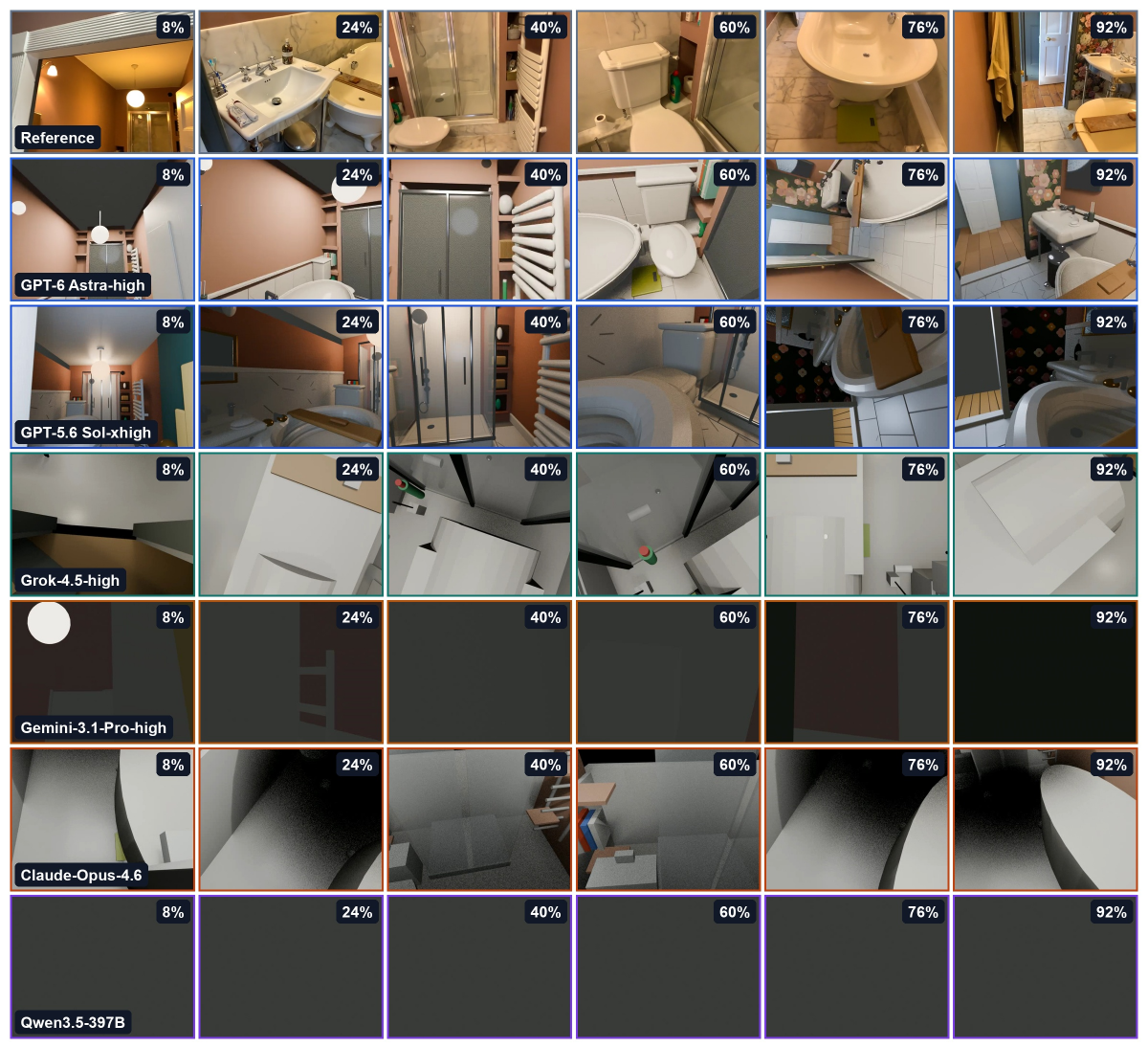}
  \caption{\textbf{Candidate 2: compact furnished room.}
  ARKitScenes scene 45260900 at six matched timestamps.}
  \label{fig:qualitative_candidate_02}
\end{figure}

\begin{figure}[H]
  \centering
  \includegraphics[width=0.80\linewidth]{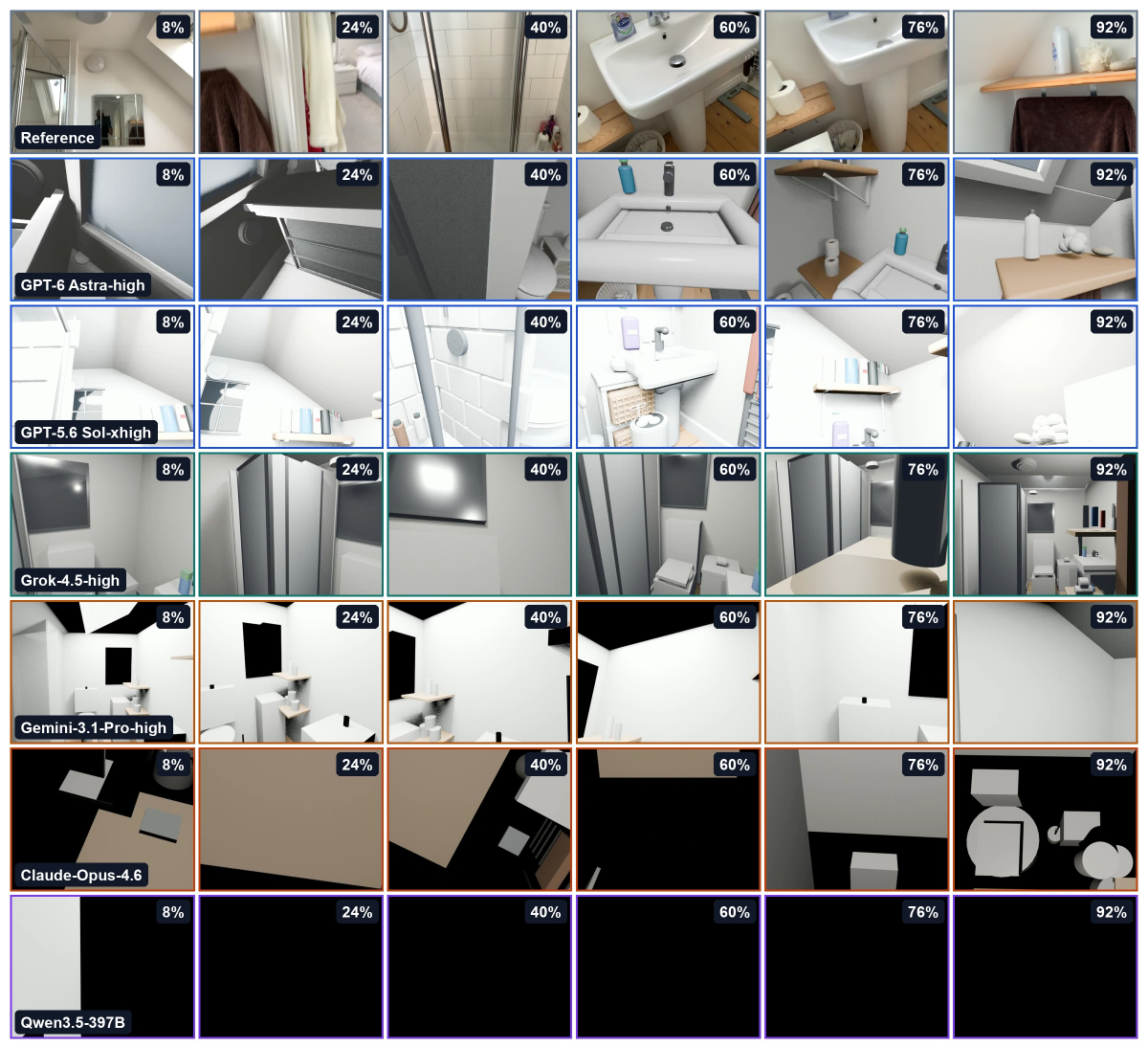}
  \caption{\textbf{Candidate 3: bathroom fixtures.}
  ARKitScenes scene 45261182 at six matched timestamps.}
  \label{fig:qualitative_candidate_03}
\end{figure}

\begin{figure}[H]
  \centering
  \includegraphics[width=0.80\linewidth]{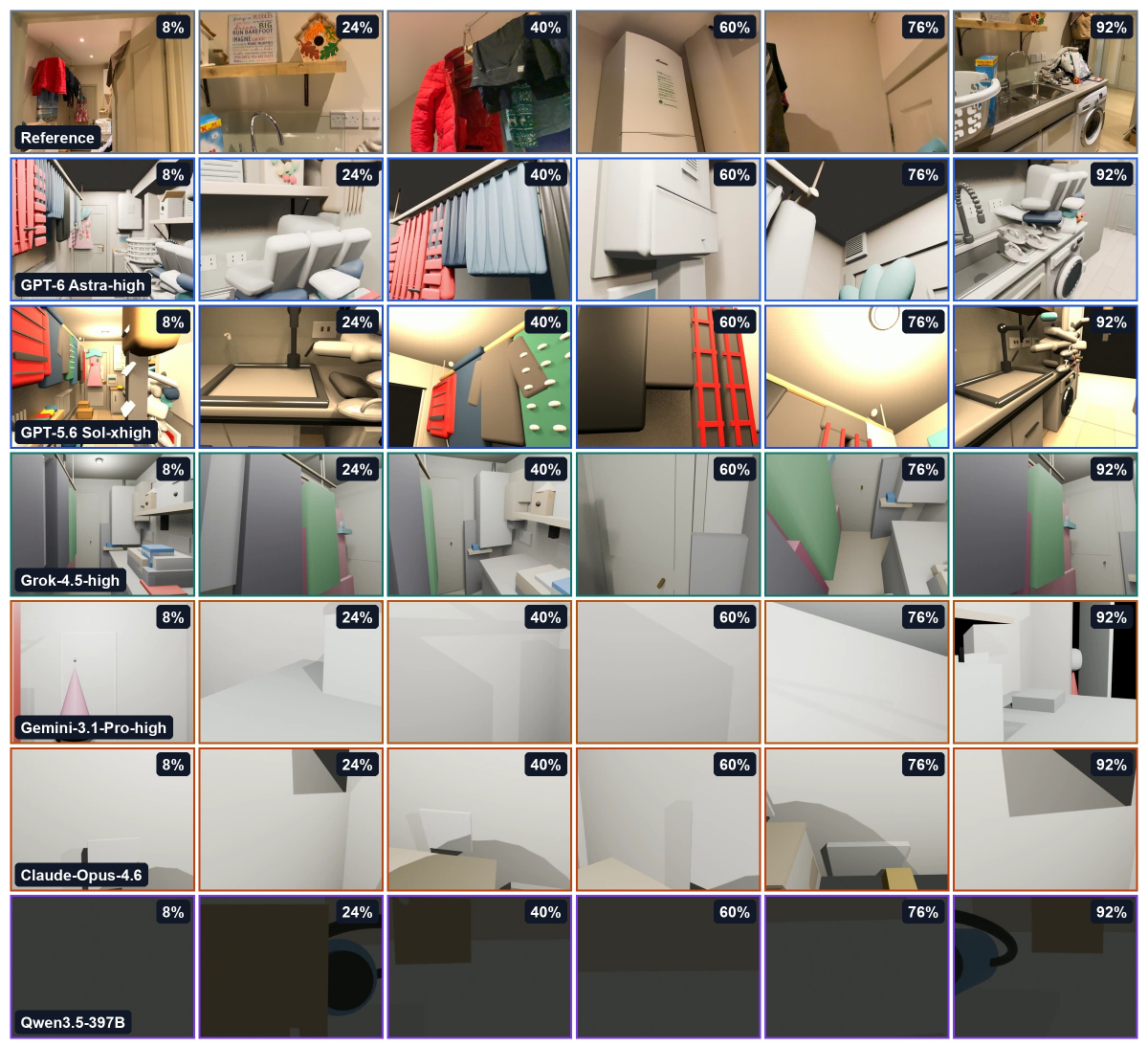}
  \caption{\textbf{Candidate 4: cluttered storage room.}
  ARKitScenes scene 42898817 at six matched timestamps.}
  \label{fig:qualitative_candidate_04}
\end{figure}

\begin{figure}[H]
  \centering
  \includegraphics[width=0.80\linewidth]{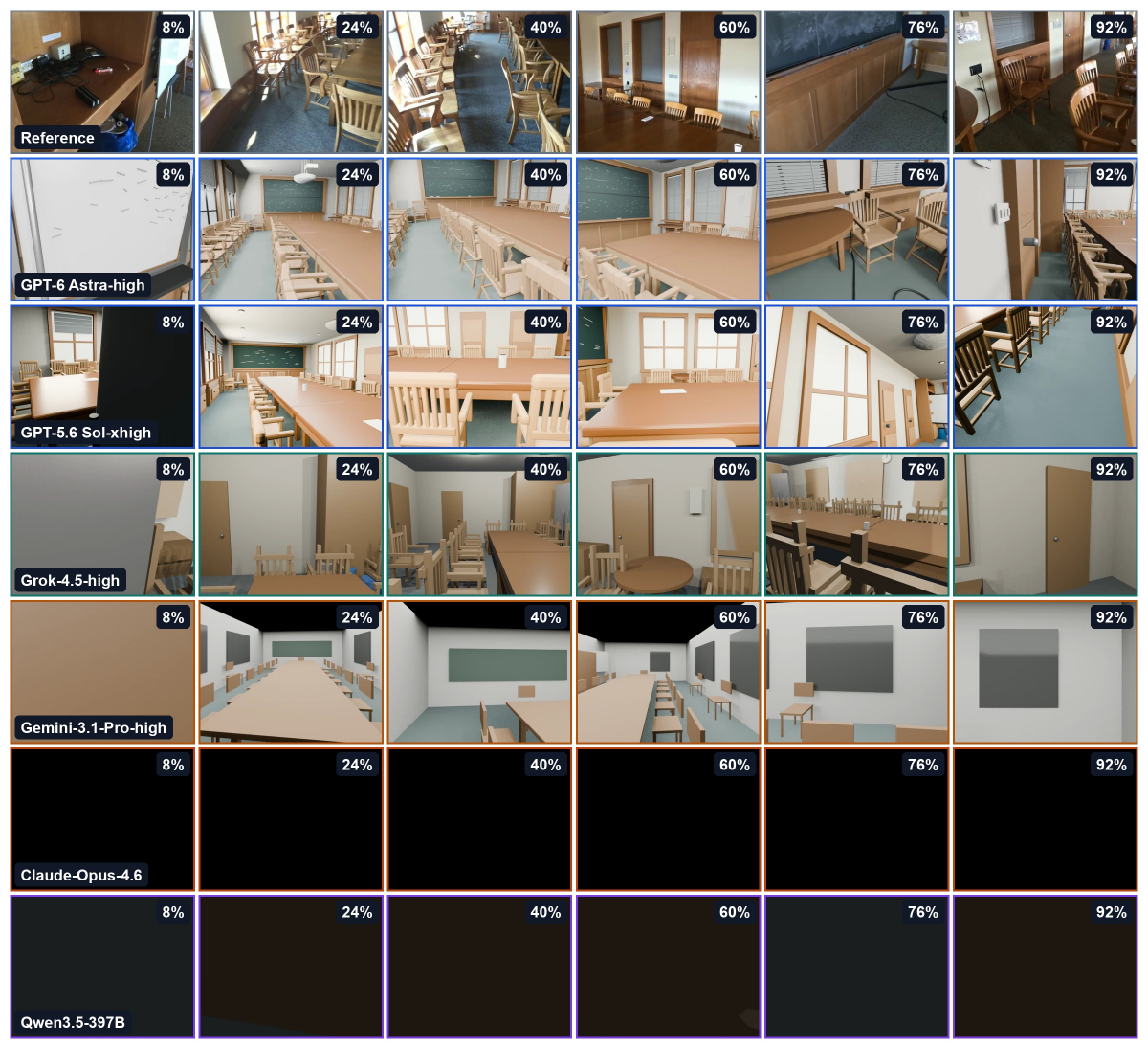}
  \caption{\textbf{Candidate 5: repeated-instance classroom.}
  ScanNet scene scene0500\_00 at six matched timestamps.}
  \label{fig:qualitative_candidate_05}
\end{figure}

\begin{figure}[H]
  \centering
  \includegraphics[width=0.80\linewidth]{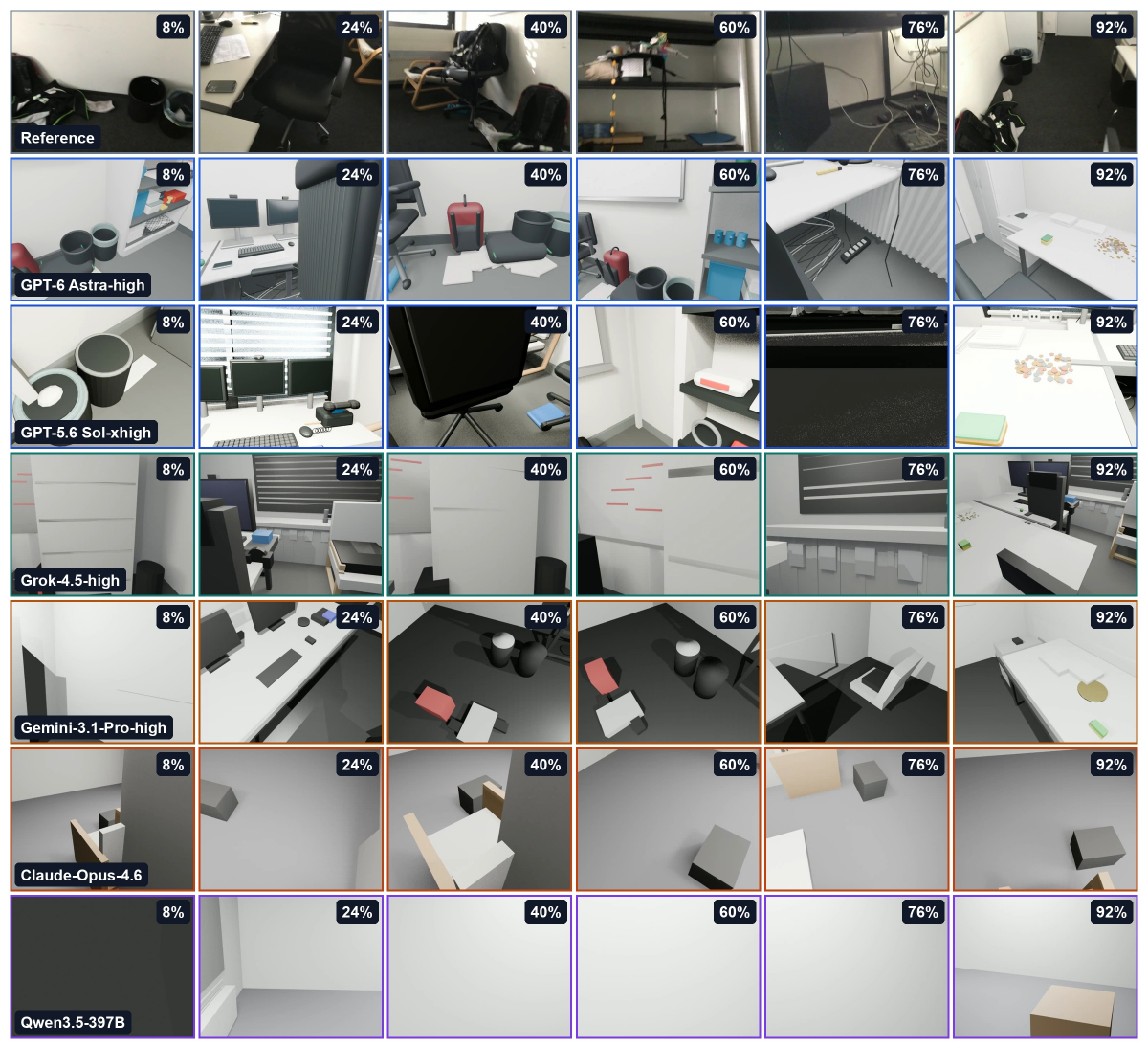}
  \caption{\textbf{Candidate 6: cluttered desktop.}
  ScanNet scene scene0700\_02 at six matched timestamps.}
  \label{fig:qualitative_candidate_06}
\end{figure}

\begin{figure}[H]
  \centering
  \includegraphics[width=0.80\linewidth]{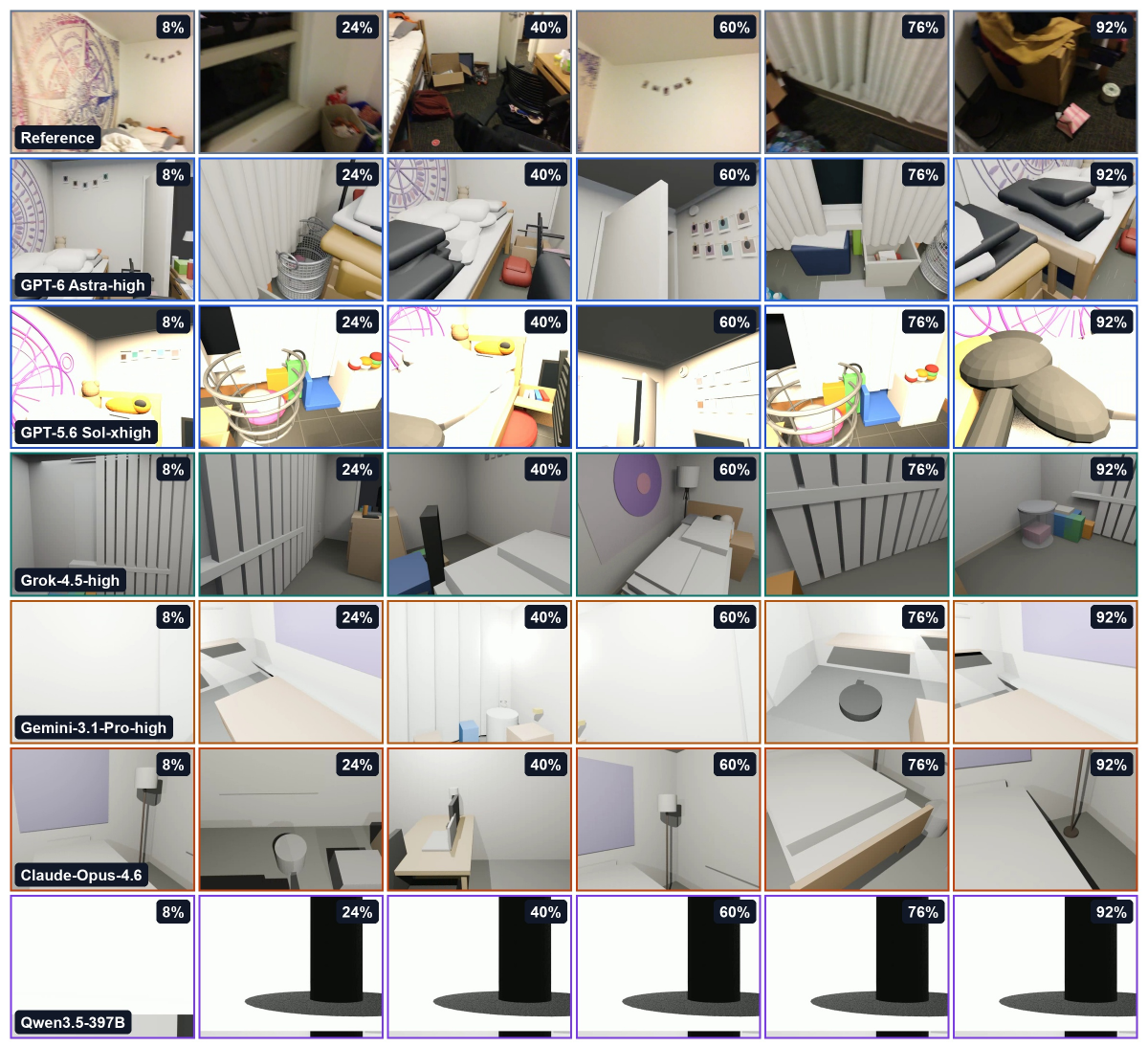}
  \caption{\textbf{Candidate 7: bedroom workspace.}
  ScanNet scene scene0695\_00 at six matched timestamps.}
  \label{fig:qualitative_candidate_07}
\end{figure}

\begin{figure}[H]
  \centering
  \includegraphics[width=0.80\linewidth]{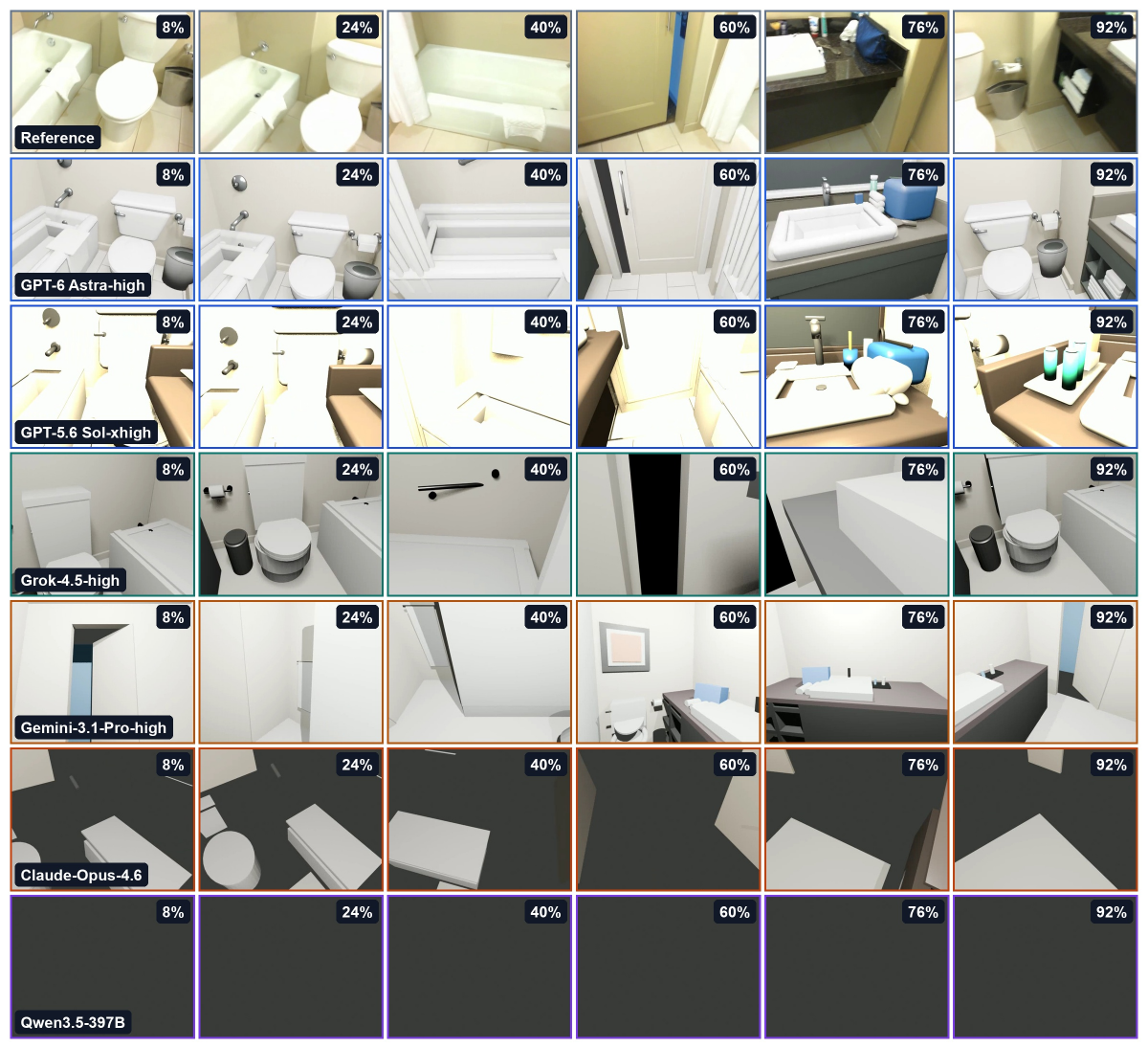}
  \caption{\textbf{Candidate 8: bathroom counter.}
  ScanNet scene scene0664\_02 at six matched timestamps.}
  \label{fig:qualitative_candidate_08}
\end{figure}

\begin{figure}[H]
  \centering
  \includegraphics[width=0.80\linewidth]{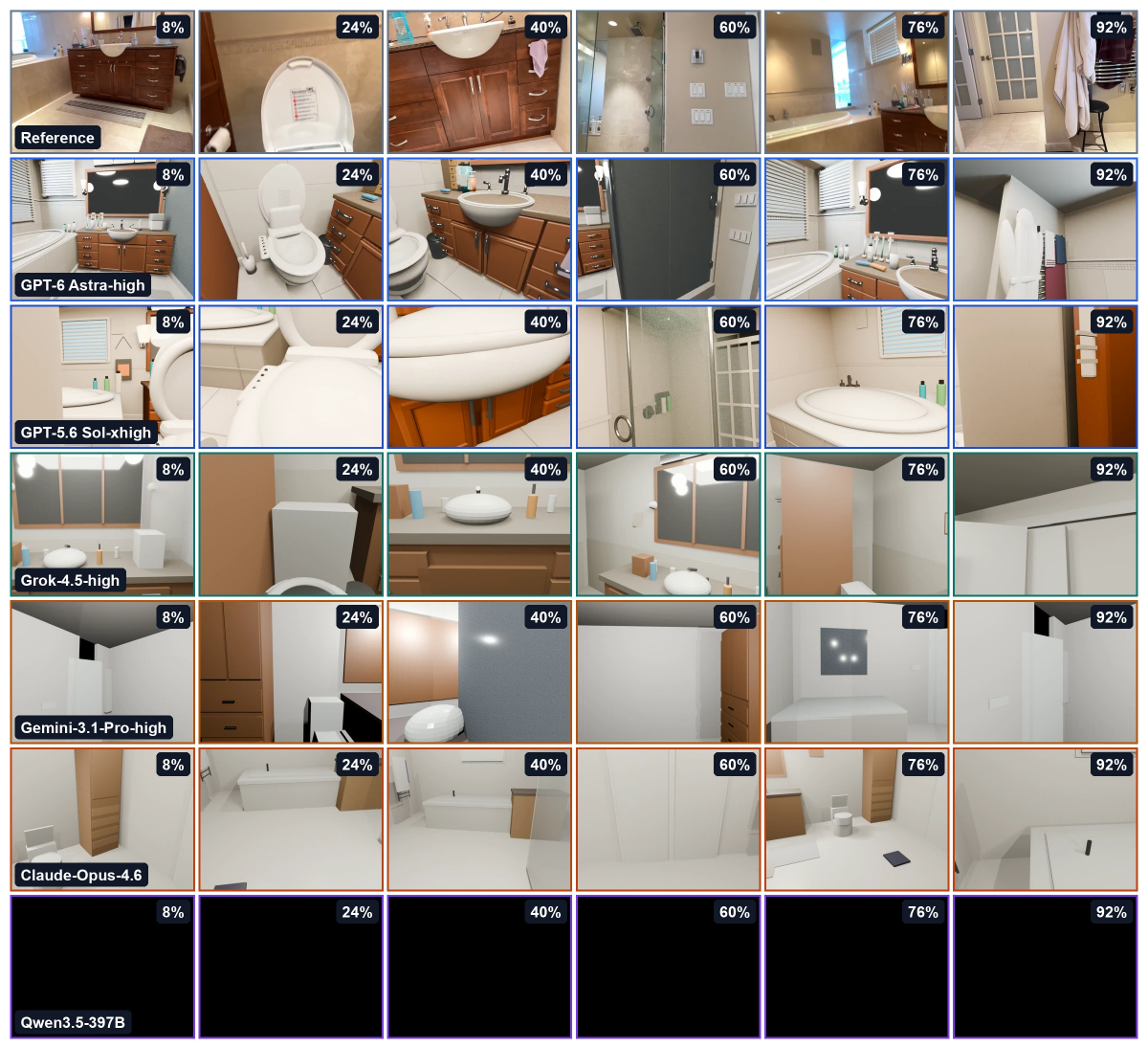}
  \caption{\textbf{Candidate 9: bathroom and doorway.}
  ScanNet++ scene e7af285f7d at six matched timestamps.}
  \label{fig:qualitative_candidate_09}
\end{figure}

\begin{figure}[H]
  \centering
  \includegraphics[width=0.80\linewidth]{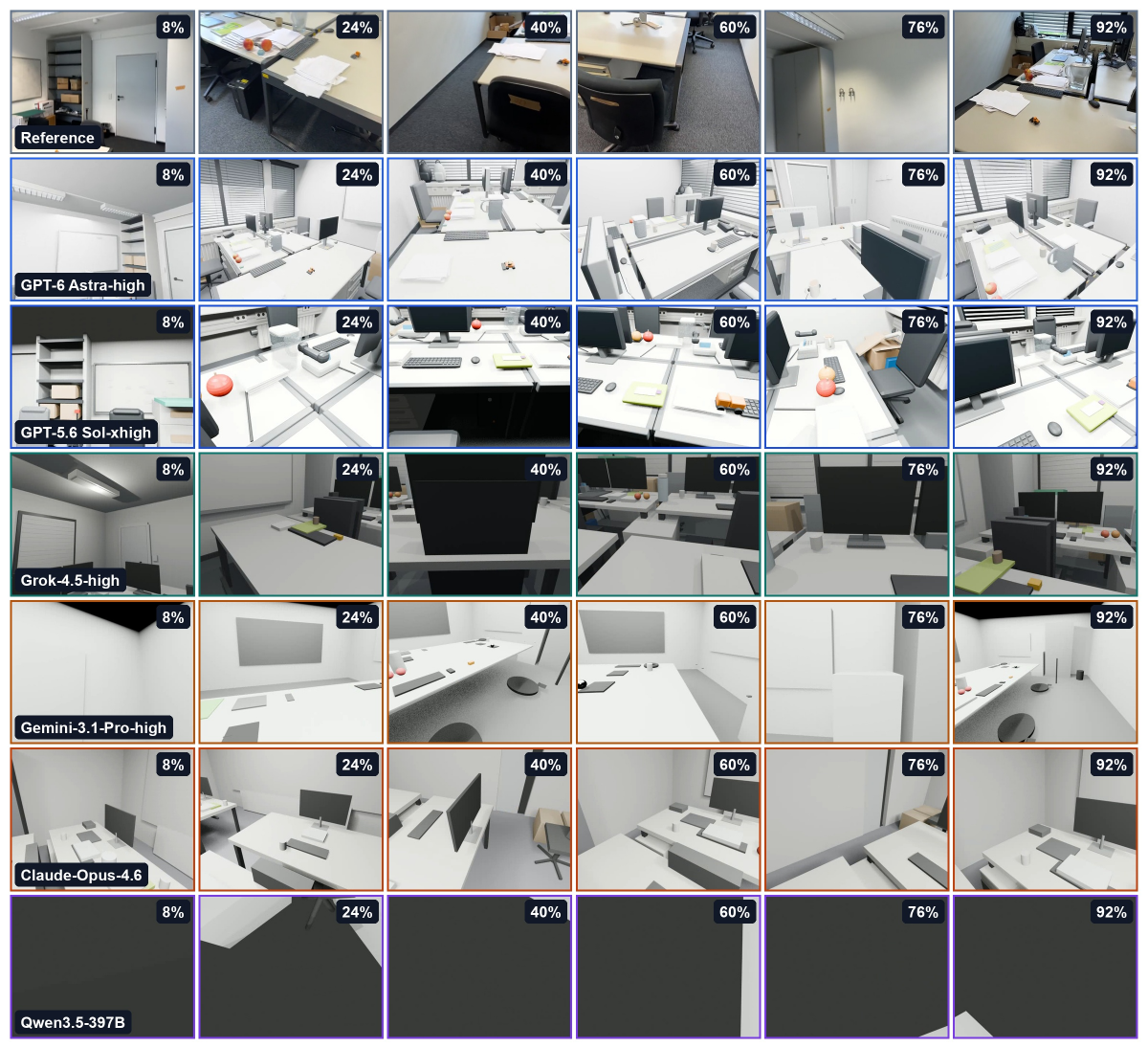}
  \caption{\textbf{Candidate 10: shared office.}
  ScanNet++ scene acd95847c5 at six matched timestamps.}
  \label{fig:qualitative_candidate_10}
\end{figure}

\begin{figure}[H]
  \centering
  \includegraphics[width=0.80\linewidth]{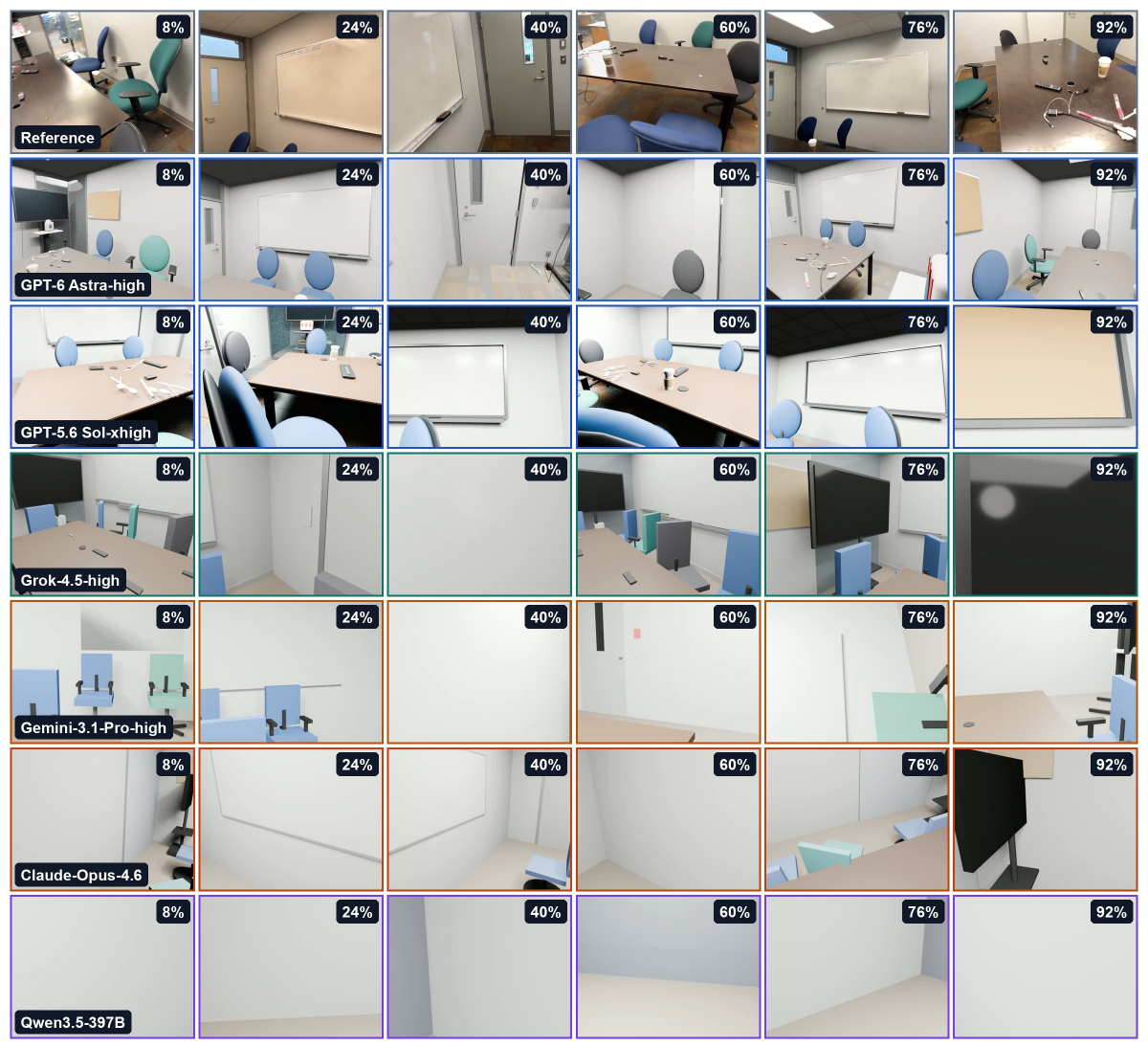}
  \caption{\textbf{Candidate 11: meeting room.}
  ScanNet++ scene 5748ce6f01 at six matched timestamps.}
  \label{fig:qualitative_candidate_11}
\end{figure}

\begin{figure}[H]
  \centering
  \includegraphics[width=\linewidth]{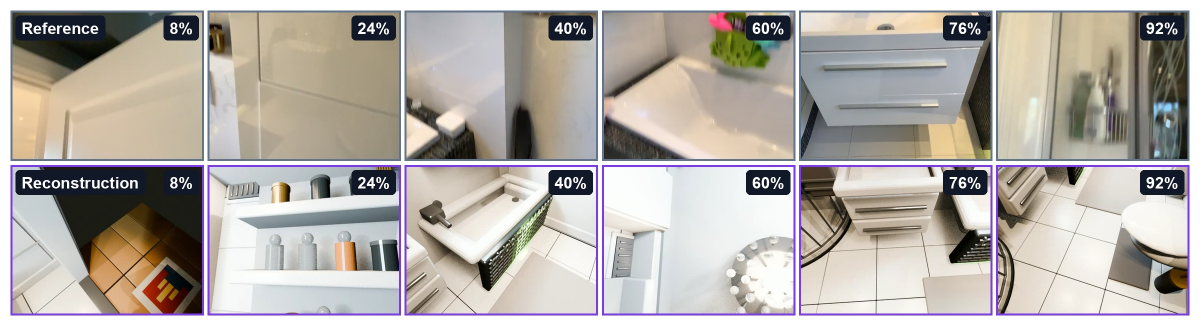}\\[-1pt]
  {\footnotesize (a) Scene 47429912: size and room-scale retention}\\[4pt]
  \includegraphics[width=\linewidth]{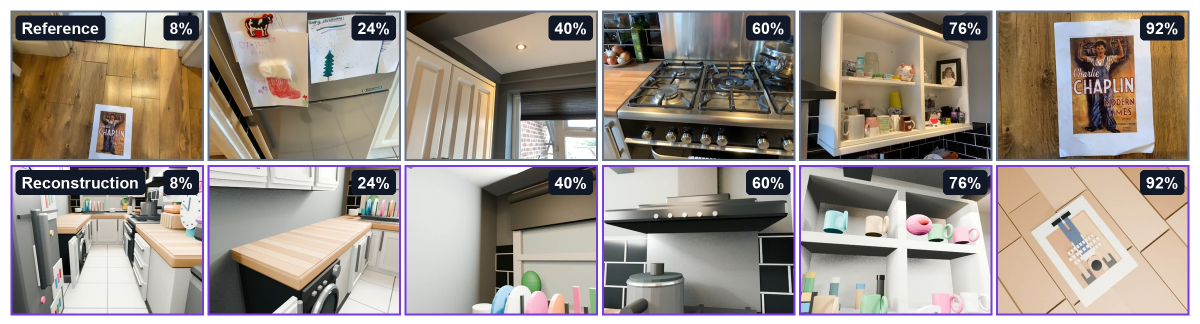}\\[-1pt]
  {\footnotesize (b) Scene 42446049: direction and route retention}
  \caption{\textbf{High perceptual similarity does not prevent localized
  semantic loss.}
  Panel (a) reaches 89.4 LS but retains neither of its two source-correct size
  questions.
  Panel (b) reaches 89.0 LS but misses its source-correct room-size,
  relative-direction, and route questions.}
  \label{fig:qualitative_high_ls_failures}
\end{figure}

\begin{figure}[H]
  \centering
  \includegraphics[width=\linewidth]{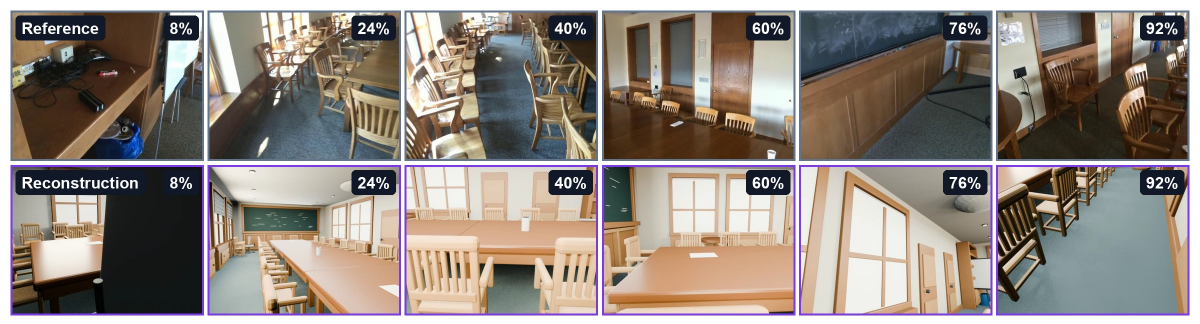}\\[-1pt]
  {\footnotesize (a) Scene scene0500\_00: object-count retention}\\[4pt]
  \includegraphics[width=\linewidth]{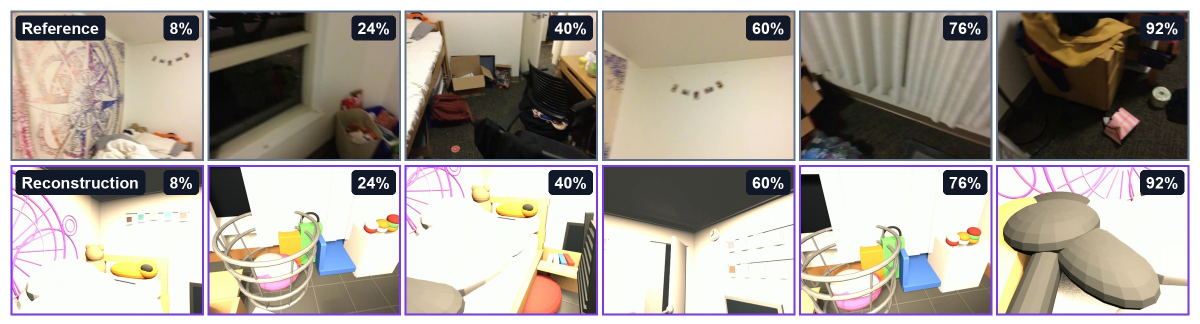}\\[-1pt]
  {\footnotesize (b) Scene scene0695\_00: appearance-order retention}
  \caption{\textbf{Inventory and temporal coverage remain distinct
  failure modes.}
  Panel (a) reaches 84.2 LS but misses its source-correct object-count question.
  Panel (b) reaches 81.8 LS but retains none of three source-correct
  appearance-order questions and misses its route question.}
  \label{fig:qualitative_inventory_order}
\end{figure}

\paragraph{Metric-complementarity cases.}
The single-configuration strips in
Figures~\ref{fig:qualitative_high_ls_failures} and
\ref{fig:qualitative_inventory_order} provide focused
source-versus-reconstruction views of four localized semantic failures.
They illustrate metric complementarity and do not estimate failure
prevalence.

\end{document}